%% file: acl_latex.tex
\documentclass[11pt]{article}

\usepackage{acl}

\usepackage{times}
\usepackage{latexsym}

\usepackage[T1]{fontenc}

\usepackage[utf8]{inputenc}

\usepackage{microtype}

\usepackage{inconsolata}

\usepackage{graphicx}
\graphicspath{{../}}

\input{math_commands.tex}

\usepackage{amsthm}
\usepackage{hyperref}
\usepackage{url}
\usepackage{algorithmic}
\usepackage{algorithm}
\usepackage{graphicx} 
\usepackage{subcaption}
\usepackage{multirow}
\usepackage{listings}
\usepackage{xcolor}
\usepackage[most]{tcolorbox}

\usepackage{makecell} 
\usepackage[table]{xcolor} 
\definecolor{lightblue}{RGB}{210,235,255}

\usepackage{amsmath}         
\usepackage{tabularx}        
\usepackage{geometry}        
\usepackage[T1]{fontenc}     
\usepackage{enumitem}        
\usepackage{booktabs}        

\lstdefinestyle{pythonstyle}{
    language=Python,
    basicstyle=\ttfamily\small,
    keywordstyle=\color{blue},
    stringstyle=\color{purple},
    commentstyle=\color{green!50!black},
    showstringspaces=false,
    breaklines=true,
    frame=tb, 
    captionpos=b
}

\title{Beyond Correctness: Harmonizing Process and Outcome Rewards through RL Training}

\author{
Chenlu Ye\textsuperscript{1,2}\thanks{Email: chenluy3@illinois.edu} \quad
Zhou Yu\textsuperscript{1} \quad
Ziji Zhang\textsuperscript{1} \quad
Hao Chen\textsuperscript{1} \quad
Narayanan Sadagopan\textsuperscript{1} \AND
Jing Huang\textsuperscript{1} \quad
Tong Zhang\textsuperscript{2} \quad
Anurag Beniwal\textsuperscript{1} \\[1em]
\textsuperscript{1}Amazon \qquad \textsuperscript{2}University of Illinois Urbana-Champaign
}

\begin{document}
\maketitle
\begin{abstract}
    Reinforcement Learning with Verifiable Rewards (RLVR) improves final-answer accuracy on reasoning tasks, but it does not reliably improve reasoning quality. Because outcome rewards only assess final answers, they also reward spurious successes: flawed reasoning can still receive maximal reward when it accidentally reaches the correct outcome. This outcome reward hacking creates biased gradients, making current RLVR insufficient for learning faithful reasoning. Process Reward Models (PRMs) provide step-wise supervision, but directly optimizing PRMs or naively combining them with outcome rewards is unstable under distribution shift during RL training process. We introduce PRocess cOnsistency Filter (PROF), a data curation method that uses PRM--ORM consistency for sample selection rather than direct reward optimization. PROF keeps correct responses with strong process support and incorrect responses with weak process support while maintaining a balanced training ratio. Experiments show that PROF consistently improves both final-answer accuracy and intermediate reasoning quality over strong baselines, with less dependence on strong PRMs. Codes and training recipes are available at \url{https://github.com/amazon-science/PROF-GRPO}.
\end{abstract}

\begin{table}[t!]
\footnotesize
\centering
\begin{tcolorbox}[
  colframe=blue!75!black,
  colback=blue!5,
  colbacktitle=blue!20,
  coltitle=black,
  fonttitle=\bfseries,
  boxrule=1pt,
  arc=5pt,
]
\footnotesize
\textbf{Problem:}
There are $4$ coins of $1, 2, 3,5$ cruzeiros, which weigh $1, 2, 3,5$  grams respectively. One of them is counterfeit, differing in weight (but not known whether it is heavier or lighter). How can the counterfeit coin be identified with the minimum number of weighings using a balance scale without weights?

\vspace{1mm}\hrule\vspace{2mm}

\textbf{Step 1: Introduction.}
We need to identify a counterfeit coin with an unknown weight difference from ...

\vspace{1mm}
\textbf{Step 2 - 6: The Flawed Weighing Logic and Code.}\\
1. Weigh coins of \textbf{1 gram and 2 grams} against coins of \textbf{3 grams and 5 grams}.
2. Based on the result of the first weighing:
    - If the left side is heavier, then the counterfeit coin is either 1 gram or 3 grams.
    - If the right side is heavier, then the counterfeit coin is either 2 gram or 5 grams.
    - If both sides are equal, then the counterfeit coin is either 2 gram or 5 grams.
3. For the second weighing, ...

Let's illustrate this with code and ensure that the steps are correct ...
output: (1, 'heavier').

\textbf{Step 7, 8, 9: Summary and Final Answer.}\\
Here is a summary of the steps:
1. Weigh coins of 1 gram and 2 grams against coins of 3 grams and 5 grams...
2. For the second weighing, weigh the 1 gram coin against a known genuine coin (3 grams)...

Thus, the minimum number of weighings required to identify the counterfeit coin is \(\boxed{2}\).
\end{tcolorbox}
\caption{An example of reaching a correct final answer through flawed reasoning. The first weighing (\{1g,2g\} vs.\ \{3g,5g\}) is invalid because the two sides are not equal in nominal weight, so all subsequent deductions are unsupported even though the final answer happens to be correct.}
\label{tab:flawed_reasoning_correct_answer}
\end{table}

\section{Introduction}

\begin{figure}[t]
    \centering
    \includegraphics[width=0.5\linewidth]{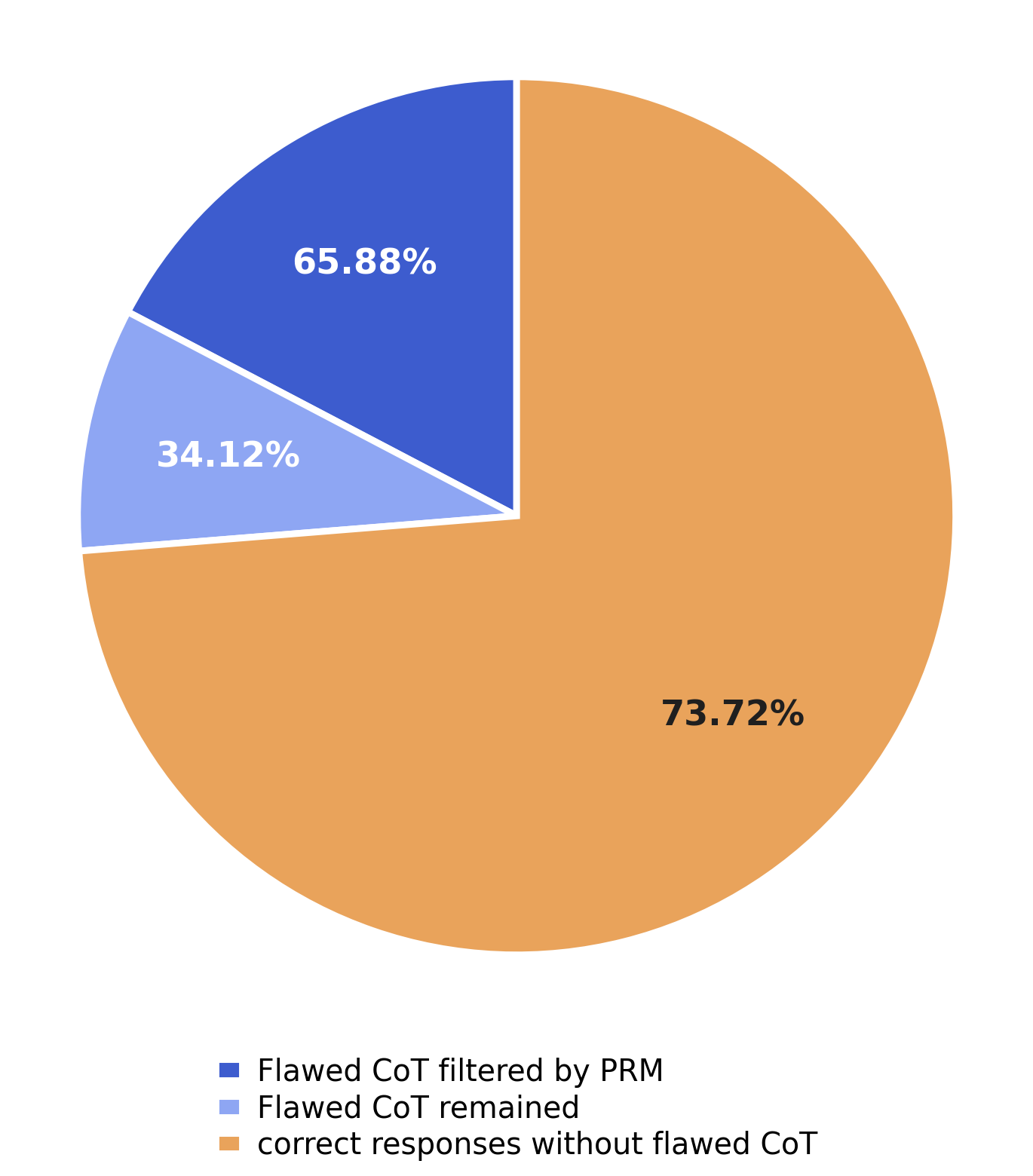}
    \caption{Within correct responses that contain flawed reasoning, PROF filters $65.88\%$ and leaves $34.12\%$.}
    \label{fig:false_positive_rate}
\end{figure}

Verifiable rewards have attracted substantial attention because they can reliably improve performance on reasoning tasks with easily verifiable outcomes, such as mathematical and coding problems \citep{cobbe2021training,jaech2024openai,shao2024deepseekmath,xiong2025self}. However, success on these tasks is usually measured only by the final answer, while in many applications we also care about the quality of the reasoning process itself, especially its faithfulness, validity, and interpretability. Throughout this paper, we use \textit{reasoning quality} as an umbrella term for these process-level properties. Optimizing the verifier is therefore not the same as optimizing reasoning quality. Because verifiers only assess final outcomes, Outcome Reward Models (ORMs) are too sparse and coarse to distinguish flawed reasoning within correct answers or valid reasoning within incorrect answers. For instance, the training example in Table~\ref{tab:flawed_reasoning_correct_answer} has fundamentally invalid reasoning but still arrives at the correct answer. To theoretically analyze this challenge, we define a latent state variable $z$, where $z=1$ denotes a valid intermediate reasoning process (i.e., no error). Let $\alpha_\pi = \mathbb{P}(z=1|\pi)$ represent the probability of generating valid reasoning. Given that an incorrect process ($z=0$) may coincidentally yield a correct answer ($r=1$) with a small probability $\epsilon$ (i.e., $\mathbb{P}(r=1|z=0) \le \epsilon$), the expected reward can be decomposed as:
\begingroup\footnotesize
\begin{equation*}
\begin{aligned}
\mathbb{E}_{\pi}[r] &= \sum_{z \in {0,1}} P(r=1 | z) P(z | \pi) \approx \alpha_\pi + \epsilon(1-\alpha_\pi)\\
&= (1-\epsilon)\alpha_\pi + \epsilon.
\end{aligned}
\end{equation*}
\endgroup
While the ideal objective is to maximize $\alpha_\pi$, the term $\epsilon(1-\alpha_\pi)$ introduces gains from spurious successes. During training, samples where $r=1$ despite $z=0$ generate biased gradients that inadvertently reinforce flawed reasoning paths, allowing the policy to increase outcome reward without improving the latent reasoning quality. This creates a \textit{process-outcome mismatch}: final-answer correctness no longer reliably reflects reasoning quality, especially the faithfulness of the underlying reasoning process. We refer to the resulting optimization failure as \textit{outcome reward hacking}: the model is rewarded for exploiting weaknesses in outcome-only supervision rather than for producing faithful reasoning. This misalignment leads to unfaithful reasoning, a limitation increasingly observed in recent studies \citep{baker2025monitoring,chen2025reasoning}. Consequently, relying solely on final answer accuracy is insufficient; ensuring reasoning quality, faithfulness, and interpretability in Chain of Thought (CoT) is crucial for the safety and practical utility of LLMs \citep{zhu2025chain,lyu2023faithful,yeo2024interpretable}.
To empirically support this process-outcome mismatch and the resulting reasoning-quality gap, we later analyze 2k samples from Qwen2.5-Math-7B and find that $26.28\%$ of correct responses still contain flawed reasoning, as judged by Claude. Within this flawed-correct subset, PROF identifies and filters $65.88\%$ (Figure~\ref{fig:false_positive_rate}).

This process-outcome mismatch shows that current RLVR alone cannot solve the reasoning-quality gap: outcome rewards are necessary for verifiability, but insufficient for supervising how the answer is reached. This has motivated a flurry of recent work on training Process Reward Models (PRMs) and using them in RL training \citep{lightman2023let,zhang2025lessons,zou2025reasonflux}, since PRMs provide dense and fine-grained feedback over intermediate reasoning processes. In other words, if we want to optimize reasoning quality rather than final correctness alone, some form of PRM-style process supervision is necessary.
However, directly using PRMs as rewards introduces a second failure mode. Although these PRMs achieve excellent performance on PRM benchmarks, directly combining PRM and ORM in the reward function can lead to reward hacking. Notably, since PRMs are often trained offline, applying them to online training suffers from distribution shift. Especially in boundary cases where the policy encounters difficult problems and produces rarely seen responses, PRMs often fail to judge them correctly, thus leading to severe reward hacking when they are used as explicit reward signals during RL training \citep{michaud2020understanding,tien2022causal}. Even when some works \citep{zha2025rl,cui2025process} attempt to co-train the policy and PRMs online, they can only train PRMs in implicit ways that lack accurate process scores, such as implicit generative rewards or alignment between process rewards and outcomes. Therefore, instead of training another PRM for a specific dataset or base model, we focus on \textbf{how to robustly integrate a pre-trained PRM into online training}, i.e., how to harmonize accurate but coarse-grained ORMs with fine-grained but noisy Process Reward Models (PRMs) in Reinforcement Learning (RL).

We develop a \textbf{PRocess cOnsistency Filter (PROF)} framework, an online data curation strategy based on process-outcome consistency. PROF oversamples responses at training time and then ranks and filters them by PRM--ORM consistency. Specifically, it removes samples where the process and outcome signals conflict, such as correct responses derived from flawed reasoning or incorrect responses that contain sound reasoning steps. By using PRMs for filtering rather than as direct optimization targets, PROF injects process supervision into RLVR while avoiding the instability of explicit PRM reward maximization. Furthermore, because correct and incorrect responses have different consistency distributions, we rank each group separately to maintain a balanced training ratio. PROF is a modular framework that can be combined with RL algorithms like Group Relative Policy Optimization (GRPO) for online training.

\begin{figure}[t]
    \centering
    \begin{subfigure}[t]{0.46\textwidth}
        \centering
        \includegraphics[width=\textwidth]{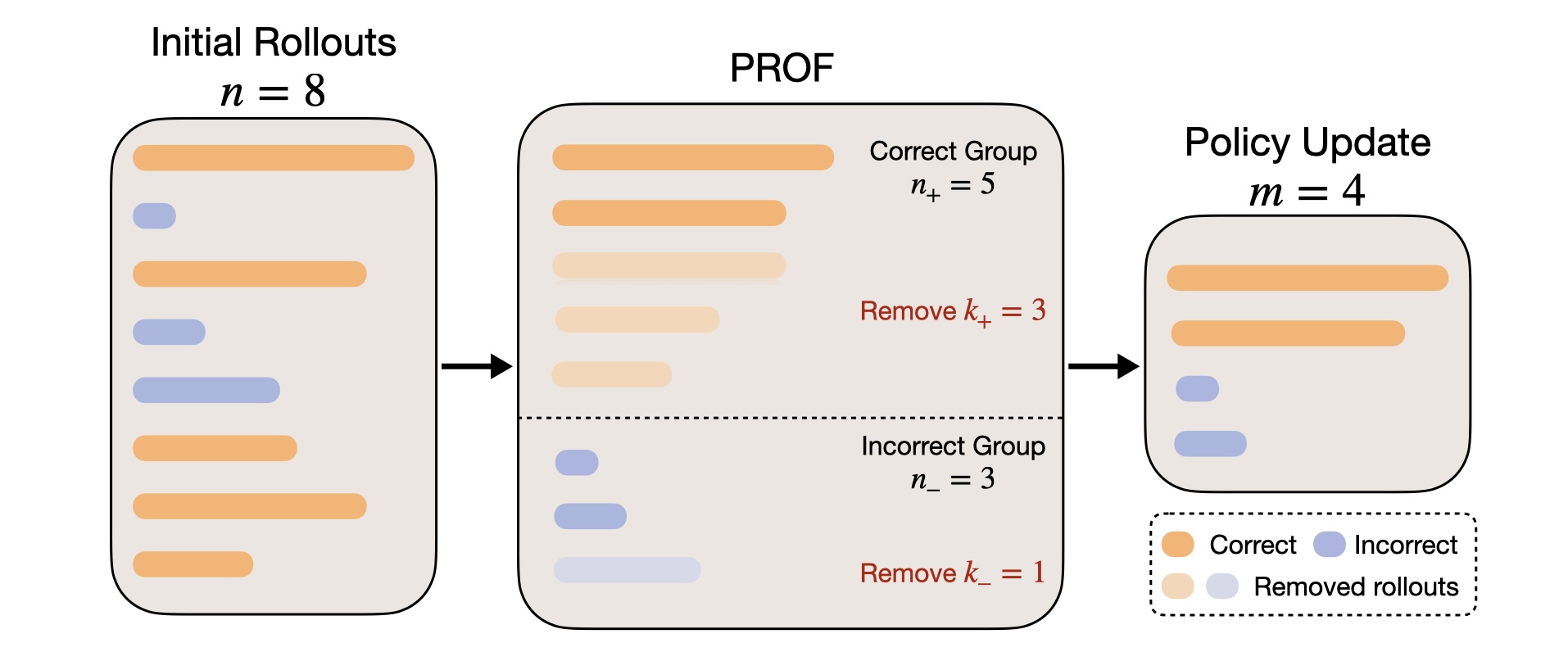}
    \end{subfigure}
    \caption{Visualization of Algorithm \ref{alg}. Rectangle length denotes trajectory-level process score. PROF ranks correct and incorrect groups separately by PRM--ORM consistency: in the correct group, higher-consistency samples are kept; in the incorrect group, lower-consistency (or random) samples are kept to maintain a balanced correct/incorrect ratio.}
    \label{fig:alg_example}
\end{figure}
We conduct extensive experiments to validate the improvement of PROF on both outcome accuracy and reasoning quality using both Qwen \citep{yang2024qwen2} and LLaMA \citep{dubey2024llama} models. To summarize, we highlight our key contributions as follows:
\begin{itemize}
    \item We identify a fundamental reasoning-quality gap in current RLVR. Because outcome-only rewards can reward spurious successes, current RLVR can improve final-answer accuracy without reliably improving faithful reasoning, a failure mode we characterize as \textit{outcome reward hacking}. We support this process-outcome mismatch with both theoretical analysis and empirical evidence.
    \item We propose \textbf{PROF}, a consistency-based data curation framework that robustly injects PRM supervision into RLVR. Rather than directly optimizing PRM scores or naively blending PRM and ORM rewards, PROF uses PRM--ORM consistency for ranking and filtering, allowing it to remove conflicting trajectories while maintaining a balanced correct/incorrect training ratio.
    \item Extensive experiments and ablations on both Qwen and LLaMA models show that PROF consistently improves both final-answer accuracy and intermediate reasoning quality over strong baselines, with less dependence on strong PRMs. Under matched compute cost and matched rollout-group size, PROF still achieves larger gains by almost $2\%$. We further demonstrate robustness to different off-the-shelf PRMs, generality beyond GRPO, and the importance of filtering correct and incorrect responses separately.
\end{itemize}

\section{Related Work}

\paragraph{Reasoning-Quality Gaps and Faithfulness of Chain-of-Thought.} A growing literature documents \textit{process-outcome mismatch} in language models: final-answer correctness can diverge substantially from reasoning quality, especially from the faithfulness of a model's verbalized reasoning. \citet{turpin2023language} show that CoT explanations can omit biasing features and rationalize incorrect predictions, while \citet{lyu2023faithful} argue that standard CoT does not guarantee a faithful explanation of how the answer is produced. Subsequent work measures this reasoning-quality gap more directly: \citet{nguyen2024direct} report a significant disparity between answer accuracy and CoT faithfulness in multi-hop question answering, and \citet{paul2024making} use causal mediation analysis to show that LLMs do not reliably use their generated intermediate steps when producing the final answer. Beyond faithfulness, \citet{yeo2024interpretable} advocate evaluating reasoning explanations along multiple axes including robustness and utility, and \citet{jacovi2024chain} show that even dedicated verifiers struggle to detect logical errors and contradictions inside reasoning chains. Recent monitoring work extends these concerns to reasoning models themselves: \citet{baker2025monitoring,chen2025reasoning} show that model-generated reasoning often fails to transparently reveal the cues or considerations that drive behavior. Relative to this line of work, we focus on the RLVR setting and study how process supervision can reduce process-outcome mismatch during online training by filtering trajectories whose final outcomes and reasoning quality are inconsistent.

\paragraph{Process-Supervised Reward Models for Fine-Grained Feedback.} RLHF focuses on \textit{trajectory-level comparison} under the Bradley-Terry model. For reasoning-related tasks, \citet{yang2024qwen2} uses final-answer correctness to construct preference pairs and trains Bradley-Terry reward models for mathematical reasoning. A more widely used approach, termed Outcome Reward Models (ORMs), trains a classifier to predict whether the final answer is correct based on the reasoning history. However, \citet{lightman2023let} show that Process-Supervised Reward Models (PRMs), which evaluate each intermediate step of a reasoning chain, significantly outperform ORMs, especially for data selection tasks such as best-of-n sampling \citep{lightman2023let}. Their approach, however, requires human annotators to label each intermediate step. \citet{wang2023math} proposes using Monte-Carlo estimation of the Q value to determine labels automatically. Many follow-up works improve PRMs through generative reward modeling, advanced training techniques such as RL, and refined engineering practices \citep{xiong2024implementation,zhang2025lessons,khalifa2025process,zhao2025genprmscalingtesttimecompute,xiong2025stepwiserstepwisegenerativejudges}. Our work does not focus on improving PRMs themselves; instead, we use PRMs to supervise the intermediate steps of CoT trajectories for data filtering. We mainly use Qwen2.5-Math-PRM-7B from \citet{zhang2025lessons} because it is trained on the Qwen distribution and achieves strong performance on ProcessBench \citep{zheng2024processbench}.

\paragraph{Sample Filtering in Reinforcement Learning for LLM.} A key challenge in applying reinforcement learning to LLM applications is the imperfection of reward signals. These signals stem from a learned reward model, such as Reinforcement Learning from Human Feedback (RLHF), or are sparse, delivered only at the end of a trajectory (e.g. RLVR). In RLHF, the reward model is trained on human-annotated pairwise comparisons, typically using a Bradley-Terry model \citep{bradley1952rank}. Due to inherent human disagreement and finite training data, the model develops shortcuts that RL algorithms can exploit \citep{lin2023mitigating, eisenstein2023helping} to chase for a fake high reward. Consequently, these rewards may not fully align with the underlying intended goals, leading to reward hacking.

Data filtering has proven effective in mitigating reward hacking across RL-based LLM training. In RLHF, prior work filters preference pairs by reward gap \citep{yuan2024self,dong2024rlhf,xiong2024building,zhang2024policy} or combines reward with response length \citep{kim2024m,yu2025rip} to retain samples that are more reliable under the learned reward model.

Filtering is also useful in RLVR despite the reward being available only at the final outcome. Rejection sampling fine-tuning discards incorrect trajectories and often approaches stronger RL baselines \citep{dong2023raft, chen2025bridging, xiong2025minimalist}. Other methods filter prompts by difficulty \citep{yang2024qwen2}, remove zero-gradient prompts via dynamic sampling \citep{yu2025dapo}, or over-sample and retain subsets that improve reward variance or the balance between correct and incorrect responses \citep{xiong2025minimalist,xu2025not}. In contrast to these methods, which mainly rely on coarse outcome-level signals, our approach uses process-supervised reward models (PRMs) \citep{lightman2023let} to filter trajectories based on reasoning quality at the level of intermediate steps and their consistency with ORMs.

\section{Formulation and Algorithm}
An LLM defines a policy distribution: given a prompt $x$, it assigns density $\pi(a|x)$ to each response $a$. For mathematical reasoning tasks with a binary verifiable reward, there exists a verifier mapping prompt-response pairs $(x,a)$ to a scalar reward $r^o(x,a)\in\{-1,1\}$. For each prompt, we generate a group of responses together with their verifier outcomes, denoted by $\{(a_i,r^o_i)\}_{i=1}^G$.

\begin{algorithm}[ht]
\caption{Process Consistency Filter (PROF)}
\footnotesize
\label{alg}
\begin{algorithmic}[1]
\STATE \textbf{Input:} Number of rollouts $n$, policy update size $m$, rollouts $\{a_1,\ldots,a_n\}$, outcome rewards $\{r^o_{1},\ldots,r^o_{n}\}$, step number regularization parameter $\lambda,H_\lambda>0$.

\STATE Obtain process rewards for each rollout $a_i$ with $H_i$ steps: $(r^1_i,\ldots,r^{H_i}_i)$ and compute trajectory-wise consistency 
\begin{equation}\label{eq:prm_calculate}
\begin{aligned}
    r^{\mathrm{pro}}_i=\Big[\frac{1}{H_i}\sum_{h=1}^{H_i}r^h_i - \lambda I(H_i=1 \text{~or~} H_i\ge H_\lambda)\Big] \cdot r^o_{i}.
\end{aligned}
\end{equation}
\STATE Divide rollouts into correct group $\mathcal G_+=\{a^+_1,\ldots,a^+_{n_+}\}$ with $r^o_i=1$ and incorrect group $\mathcal G_-=\{a^-_1,\ldots,a^-_{n_-}\}$ with $r^o_i=-1$, where $n_+ + n_- =n$.
\STATE Compute kept number $k_+\in[n_+],k_-\in[n_-]$ in each group such that $k_++k_-=m$ and $k_+k_-$ is maximized.
\STATE Rank $\mathcal{G}_+$ and $\mathcal{G}_-$ by $r^{\mathrm{pro}}$ separately, and for the correct group keep the samples $\mathcal{K}^+=\{a^+_i| \mathrm{rank}(r^{\mathrm{pro}}(a^+_i))\le k_+\}$.\\
For the incorrect group, we have two sub-algorithms:
\begin{itemize}
    \item {\color{teal}{PROF-POS}}: randomly pick $k_-$ samples from $\mathcal G_-$;
    \item {\color{pink}{PROF-BOTH}}: keep $\mathcal{K}^-=\{a^-_i| \mathrm{rank}(r^{\mathrm{pro}}(a^-_i))\ge n_- - k_-\}$.
\end{itemize}
\STATE \textbf{Output:} The kept trajectories $\mathcal{K}^+ \cup \mathcal{K}^-$ with final kept size $m$.
\end{algorithmic}
\end{algorithm}

\paragraph{GRPO.} \citep{shao2024deepseekmath} proposes this policy gradient algorithm that simplifies the Proximal Policy Optimization (PPO) \citep{schulman2017proximal} by only computing the advantage based on the outcome rewards in a group. Instead of maintaining and updating another value network, GRPO computes the advantage by standardizing the outcome rewards within a group: for $i=1,\ldots,n$,
\begingroup\small
\begin{equation*}
\begin{aligned}
    A_i = \frac{r^o(x, a_i) - \text{mean}\left(\{r^o(x, a_j)\}_{j=1}^n\right)}{\text{std}\left(\{r^o(x, a_j)\}_{j=1}^n\right) + \delta},
\end{aligned}
\end{equation*}
\endgroup
where $r^o(x, a_i)$ is the outcome reward for a given response and $\delta>0$ is a small constant for numerical stability. Let $a_i^t$ denote the $t$-th token and $a_i^{<t}$ denote $(a_i^1,\ldots,a_i^{t-1})$. This advantage is then incorporated into a clipped surrogate objective, which is optimized to update the policy from $\pi_{\theta_{\text{old}}}$ to $\pi_\theta$:
\begingroup\footnotesize
\begin{equation*}
\begin{aligned}
    \mathcal{J}_{\text{GRPO}}(\theta) =& \mathbb{E}_{x \sim \mathcal{D}} \bigg[ \frac{1}{n} \sum_{i=1}^{n}\sum_{t=1}^{|a_i|}\min \Big( \frac{\pi_\theta(a_i^t|x,a_i^{<t})}{\pi_{\theta_{\text{old}}}(a_i^t|x,a_i^{<t})} A_i,\\
    &\quad \text{clip} \left( \frac{\pi_\theta(a_i^t|x,a_i^{<t})}{\pi_{\theta_{\text{old}}}(a_i^t|x,a_i^{<t})}, 1-\epsilon, 1+\epsilon \right) A_i \Big) \bigg].
\end{aligned}
\end{equation*} 
\endgroup

Although this approach stabilizes the online policy optimization and is efficient, the sparse reward signal limits further improvement in intermediate reasoning quality.

\paragraph{Process Reward Model (PRM).} For a response $a$ composed of multiple reasoning steps $a=(a^1,\ldots,a^H)$, we follow previous works \citep{zheng2024processbench,zhang2025lessons,zou2025reasonflux} to use a newline as a sign for a new step. For each step $a^h$, the PRM score $r^h$ maps it, the previous steps and the prompt $(x, a^{\le h})$ to a scalar $r^h(x, a^{\le h})$, where we use the short-hand notation $a^{\le h}=(a^1,\ldots,a^h)$.

\paragraph{Our Method PROF: Process Consistency Filter Framework}
We propose PROF in Algorithm~\ref{alg} to robustly incorporate PRM--ORM consistency after the rollout phase, and also visualize it in Figure~\ref{fig:alg_example}. First, we generate $n$ samples and obtain outcome rewards. Then, we call the PRM to generate step-level rewards for each rollout and compute the trajectory-wise consistency score $r^{\mathrm{pro}}$ by taking the mean over step-level rewards and adding a step-length regularization in \eqref{eq:prm_calculate}, where $\lambda$ is the regularization parameter and $H_\lambda$ is the threshold for the penalized step number. This regularization ensures that samples with no step segments or over-long steps are discarded in the correct group. The samples are divided into two subgroups: $\mathcal{G}_+$ contains correct samples with $r^o=1$, and $\mathcal{G}_-$ contains incorrect samples with $r^o=-1$. Inspired by \citep{xu2025not}, the numbers to keep in each subgroup, $k_+$ and $k_-$, are chosen to maximize the outcome-reward variance of the final kept samples $k_+k_-/(k_++k_-)^2$. Since $k_++k_-=m$ is fixed, $k_+k_-=k_+(m-k_+)$ should be maximized, and the maximum is attained when $k_+$ is closest to $m/2$ under the constraint $k_+\le n_+, k_-\le n_-$. This implies that the ratio of correct and incorrect responses should be balanced. After that, for the correct group, we use $r^{\mathrm{pro}}$ to rank and keep the top $k_+$ samples. For the incorrect group, PROF-POS randomly filters samples, while PROF-BOTH uses $r^{\mathrm{pro}}$ to rank and keep the bottom $k_-$ samples.
Finally, we collect the kept $m$ trajectories for policy update.

\paragraph{False Positives Are Frequent and Filterable.}
We provide empirical evidence in Figure \ref{fig:false_positive_rate} to justify the practical motivation. On 2k samples from Qwen2.5-Math-7B, we find that $26.28\%$ of correct responses still exhibit flawed reasoning, as judged by Claude. Crucially, within this flawed-correct subset, when PROF filters the bottom half of correct responses by PRM consistency, it identifies and removes $65.88\%$ of these flawed responses. This confirms that process-outcome mismatch is a critical bottleneck and that PROF effectively filters problematic samples to improve gradient quality.

\section{Experiments}
\subsection{Setup}
We focus on mathematical reasoning tasks in this work. For online training, we use the Numina-Math prompt set \citep{numina_math_7b}, which contains nearly 860k math problems with ground-truth answers ranging from Chinese high school exercises to US and international mathematics olympiad problems. We use Qwen2.5-Math-1.5B-base and Qwen2.5-Math-7B-base \citep{yang2024qwen2} as the training base models. For the PRM, we mainly use Qwen2.5-Math-PRM-7B \citep{zhang2025lessons} to generate process rewards. We also experiment with a weaker PRM, Skywork-PRM-1.5B \citep{he_2024_16998085}, to study the robustness of PROF to PRM quality. More details are provided in Appendix~\ref{sec:additional_experimental_details}. Model performance is evaluated on five benchmarks: Math500 \citep{hendrycks2021measuring}, Minerva Math \citep{lewkowycz2022solving}, Olympiad Bench \citep{he2024olympiadbench}, AMC2023\footnote{https://huggingface.co/datasets/math-ai/amc23}, and AIME2024\footnote{https://huggingface.co/datasets/math-ai/aime24}. We mainly use average@$16$ for evaluation, i.e., accuracy averaged over $16$ responses per prompt under temperature $1.0$. The models are allowed to generate $4096$ tokens.

\subsection{Main Results}
\begin{table*}[ht]
\footnotesize
\centering
\setlength{\tabcolsep}{4pt}
\begin{tabular}{cc|cccccc}
\hline
Model & Algorithm & Math500 & Minerva Math & Olympiad Bench & AIME24 & AMC23 & Average \\
\hline
\multirow{4}{*}{\makecell[l]{Qwen2.5-Math-\\1.5B-base}} 
& Base            & 39.9 & 11.4 & 19.1 & 3.5 & 23.6 & 19.5 \\
& GRPO            & 70.3 & 29.1 & 33.0 & 9.0 & 44.5 & 37.2 \\
& Blend           & 67.6 & 27.8 & 31.1 & 7.7 & 42.5 & 35.3 \\
& \cellcolor{lightblue}PROF-POS & \cellcolor{lightblue}72.6 & \cellcolor{lightblue}\textbf{31.3} & \cellcolor{lightblue}\textbf{36.1} & \cellcolor{lightblue}\textbf{10.6} & \cellcolor{lightblue}\textbf{50.3} & \cellcolor{lightblue}\textbf{40.2} \\
& \cellcolor{lightblue}PROF-BOTH & \cellcolor{lightblue}\textbf{73.2} & \cellcolor{lightblue}30.0 & \cellcolor{lightblue}\textbf{36.1} & \cellcolor{lightblue}9.6 & \cellcolor{lightblue}49.1 & \cellcolor{lightblue}39.6 \\
\hline
\multirow{4}{*}{\makecell[l]{Qwen2.5-Math-\\7B-base}} 
& Base            & 42.0 & 12.8 & 19.2 & 12.9 & 30.0 & 23.4 \\
& GRPO            & 81.6 & 37.2 & 45.5 & 20.6 & 64.4 & 49.9 \\
& Blend           & 81.7 & 36.7 & 45.0 & 15.2 & 58.0 & 47.3 \\
& \cellcolor{lightblue}PROF-POS & \cellcolor{lightblue}81.4 & \cellcolor{lightblue}36.6 & \cellcolor{lightblue}45.0 & \cellcolor{lightblue}\textbf{24.8} & \cellcolor{lightblue}64.2 & \cellcolor{lightblue}50.6 \\
& \cellcolor{lightblue}PROF-BOTH & \cellcolor{lightblue}\textbf{83.1} & \cellcolor{lightblue}\textbf{39.0} & \cellcolor{lightblue}\textbf{47.8} & \cellcolor{lightblue}17.5 & \cellcolor{lightblue}\textbf{70.9} & \cellcolor{lightblue}\textbf{51.7} \\
\hline
\multirow{4}{*}{\makecell[l]{LLaMA-3.2-\\3B-instruct t}} 
&Base                & 30.0 & 8.8 & 6.1 & 2.3 & 10.6 & 11.6 \\
&GRPO                & 50.5 & 18.8 & 17.9 & 5.0 & 25.6 & 23.6\\
&Blend     & 37.2 & 13.1 & 9.9  & 1.0 & 17.2 & 15.7\\
& \cellcolor{lightblue}PROF-POS & \cellcolor{lightblue}\textbf{52.4} & \cellcolor{lightblue}\textbf{19.5} & \cellcolor{lightblue}\textbf{19.8} & \cellcolor{lightblue}\textbf{6.7} & \cellcolor{lightblue}\textbf{28.6} & \cellcolor{lightblue}\textbf{25.4} \\
& \cellcolor{lightblue}PROF-BOTH & \cellcolor{lightblue}49.0 & \cellcolor{lightblue}18.0 & \cellcolor{lightblue}17.3 & \cellcolor{lightblue}5.4 & \cellcolor{lightblue}23.9 & \cellcolor{lightblue}22.7 \\
\hline
\end{tabular}
\caption{Performance across Math500, Minerva Math, Olympiad Bench, AMC2023, and AIME2024. Blend denotes Blend-PRM-GRPO. Reported accuracy is average@16 under temperature $1.0$, with each method tuned to its best setting.}
\label{tab:main_benchmark}
\end{table*}
We summarize our main results in Table \ref{tab:main_benchmark}, where Blend denotes a common way that mixes the PRM with outcome rewards \citep{zha2025rl,cui2025process,zou2025reasonflux}. Following \citep{zou2025reasonflux}, the PRMs are averaged over steps for each response, weighted by a parameter $\beta$, and added to outcome rewards. We use parameter $\beta=0.8$ according to Table 5 of \citep{zou2025reasonflux}. Our main findings are as follows.

\begin{figure}[t]
    \centering
    \begin{subfigure}[t]{0.48\linewidth}
        \centering
        \includegraphics[width=\linewidth]{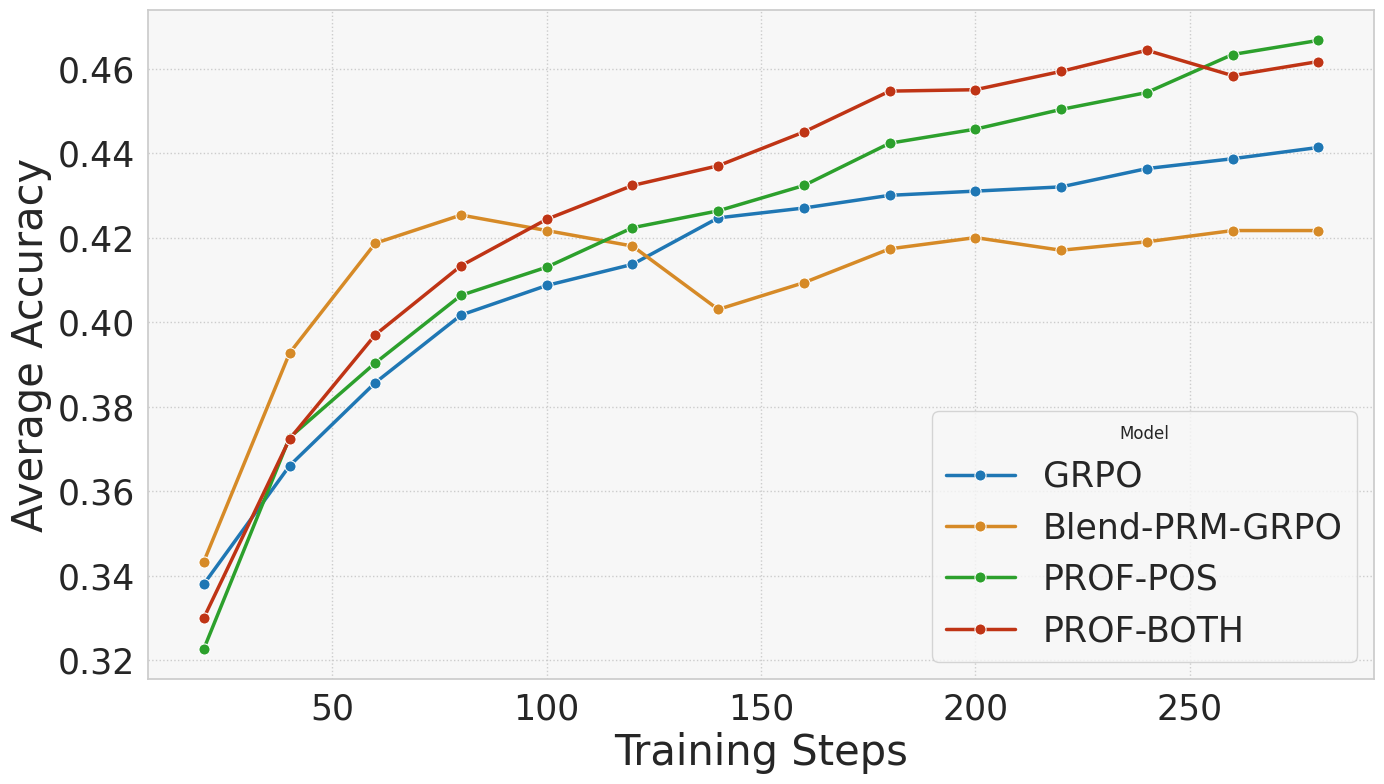}
    \end{subfigure}
    \hfill
    \begin{subfigure}[t]{0.48\linewidth}
        \centering
        \includegraphics[width=\linewidth]{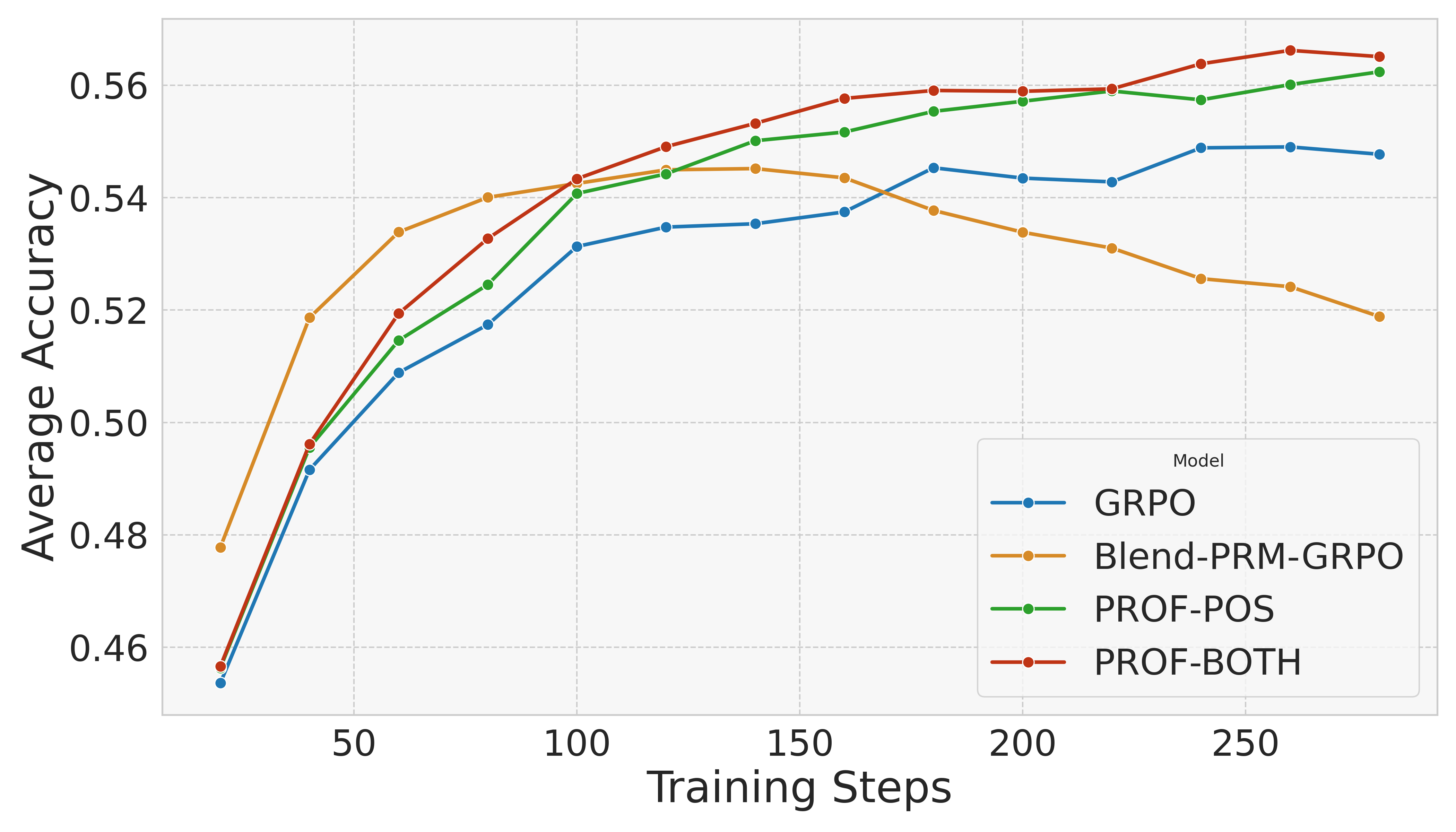}
    \end{subfigure}
    \caption{Learning dynamics from Qwen2.5-Math-1.5B-base (left) and Qwen2.5-Math-7B-base (right), compared with GRPO and Blend-PRM-GRPO. The y-axis is average@16 accuracy further averaged over Math500, Minerva Math, and Olympiad Bench.}
    \label{fig:main}
\end{figure}

\paragraph{PROF Improves Accuracy under Standard and Matched-Cost Comparisons.} 
As shown in Table~\ref{tab:main_benchmark}, our proposed methods, PROF-POS and PROF-BOTH, consistently outperform GRPO and Blend-PRM-GRPO across benchmarks and base models. For models starting from Qwen2.5-Math-1.5B-base, PROF-POS and PROF-BOTH achieve average accuracies of $40.2\%$ and $39.6\%$, surpassing the standard GRPO baseline ($37.2\%$) and Blend-PRM-GRPO ($35.3\%$). A similar trend is observed with Qwen2.5-Math-7B-base, where PROF-POS and PROF-BOTH achieve $50.6\%$ and $51.7\%$ average accuracies, significantly above GRPO's $49.9\%$ and Blend-PRM-GRPO's $47.3\%$. Moreover, for LLaMA-3.2-3B-instruct, whose policy distribution differs from the Qwen family, Blend performs even worse than GRPO, while PROF-POS still outperforms the baseline by $1.8\%$. The learning dynamics in Figure~\ref{fig:main} corroborate these findings, illustrating that PROF steadily maintains a consistent performance advantage over both GRPO and Blend-PRM-GRPO throughout training, with faster convergence and higher final accuracy than GRPO. 

\begin{figure}[t]
    \centering
    \begin{subfigure}[t]{0.49\linewidth}
        \centering
        \includegraphics[width=\linewidth]{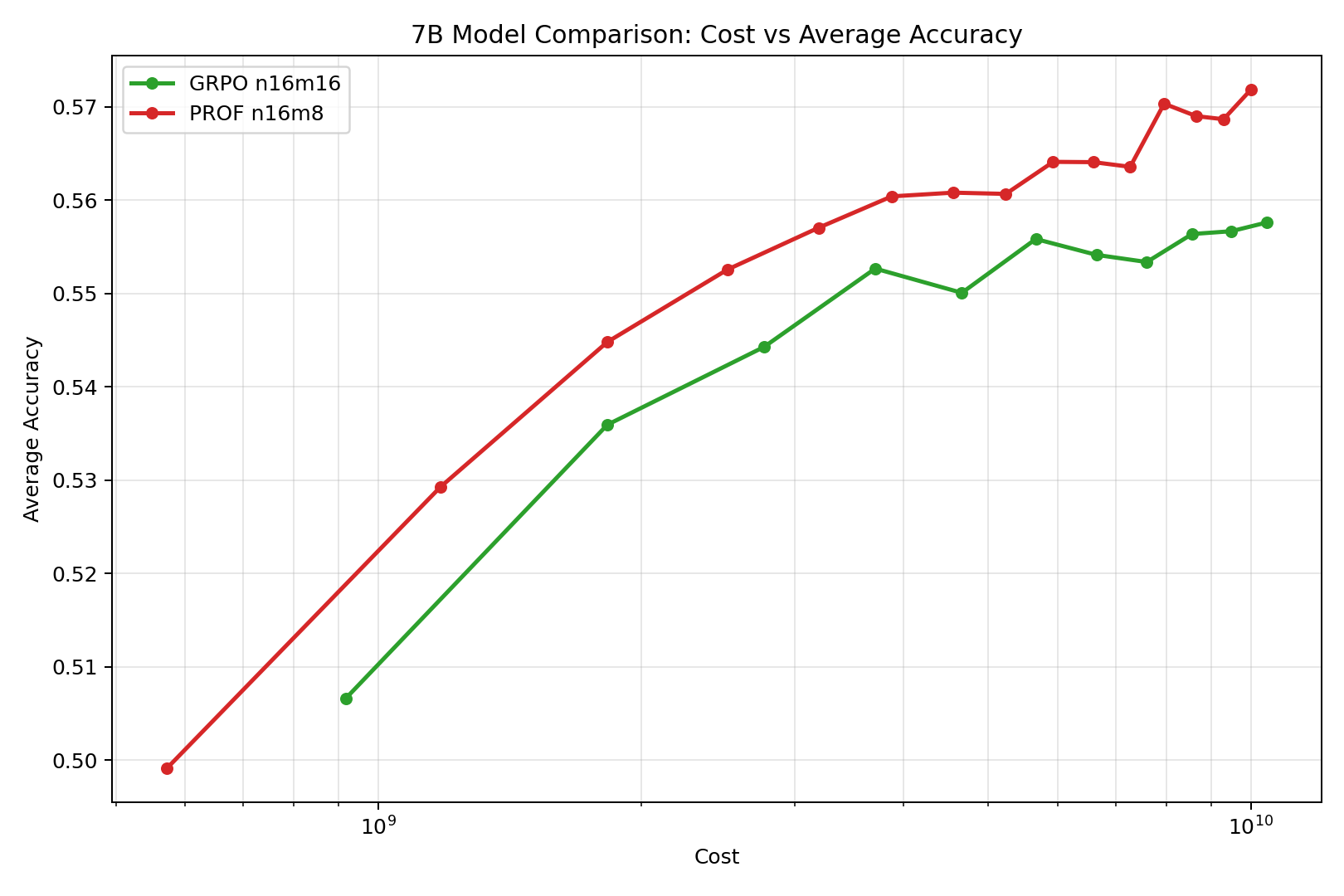}
    \end{subfigure}
    \hfill
    \begin{subfigure}[t]{0.48\linewidth}
        \centering
        \includegraphics[width=\linewidth]{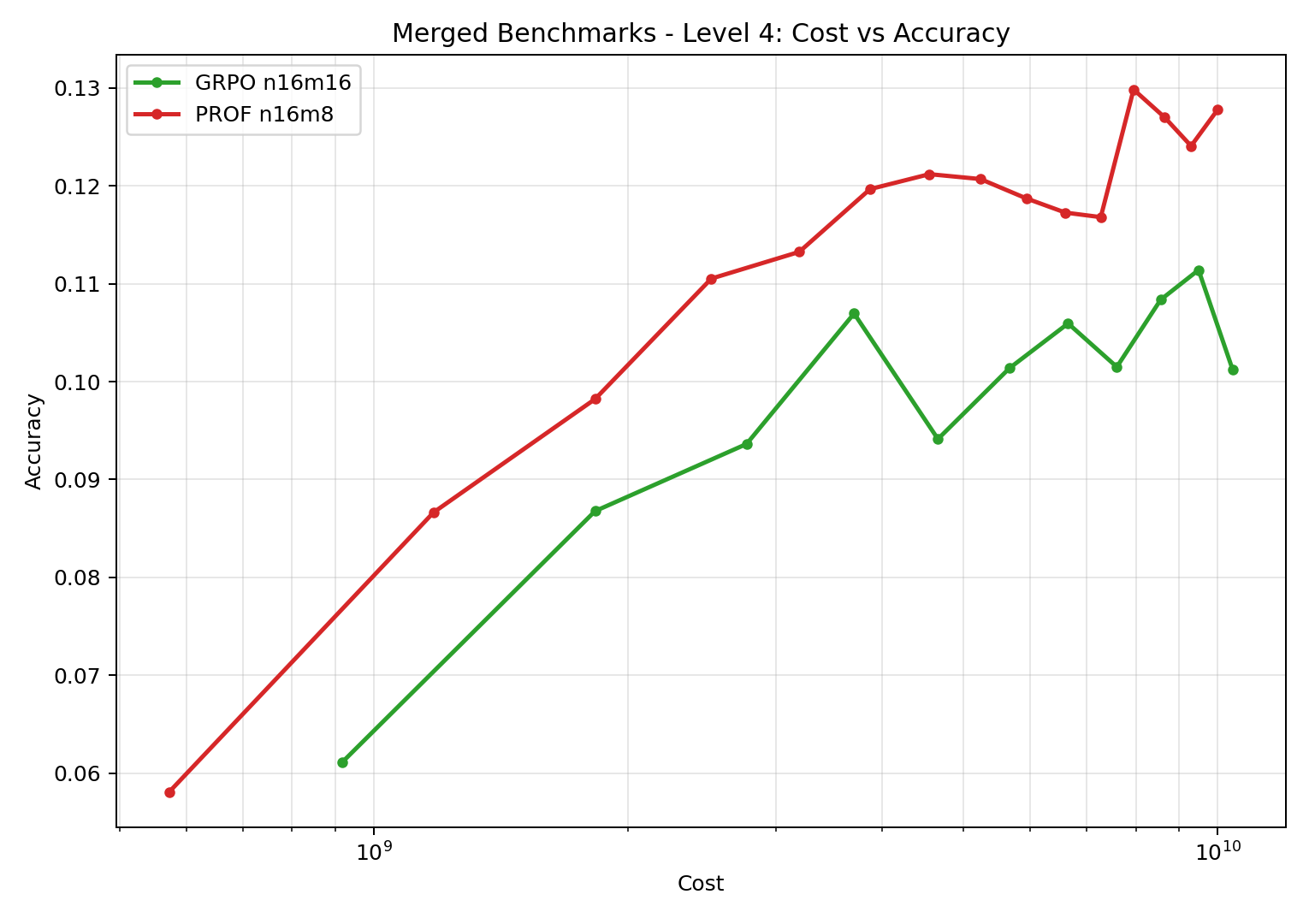}
    \end{subfigure}
    \caption{Matched-cost comparison on Qwen2.5-Math-7B-base: overall average accuracy averaged over $5$ benchmarks (left) and Level-4 hardest bucket (right). Average cost is computed as Inference $+ 3\times$ Train $+$ PRM.}
    \label{fig:main_cost}
\end{figure}

To further address efficiency and fairness concerns, we increase the rollout group size $n$ and policy update group size $m$, and compare GRPO-$n16m16$ with PROF-$n16m8$ on Qwen2.5-Math-7B-base under matched compute cost. As shown in Figure~\ref{fig:main_cost}, PROF achieves larger gains than GRPO at the same cost level. We compute average cost as \textit{Inference $+ 3\times$ Train $+$ PRM}, where the factor $3$ is a rough FLOPs proxy for training relative to a forward pass. We further aggregate all five benchmarks by base-model pass rate into four difficulty levels: Level 1 ($p>0.5$), Level 2 ($0.25<p\le0.5$), Level 3 ($0<p\le0.25$), and Level 4 ($p=0$). PROF's gain is especially pronounced on harder problems, plausibly because easier problems usually involve shorter and simpler reasoning with fewer flaws, making improvements smaller and more sensitive to PRM noise, whereas harder problems rely much more on PRM's ability to distinguish trajectory quality. Due to space constraints, Figure~\ref{fig:main_cost} visualizes only Level 4 in the main text, while matched-cost curves for Levels 1--3 are provided in Appendix Figure~\ref{fig:cost_level_123_appendix}.

\paragraph{Filtration Method is Much More Robust than Blending.}
We plot the entropy loss and response length curves of GRPO, Blend-PRM-GRPO, and PROF in Figure~\ref{fig:ent}. Blend-PRM-GRPO suffers from severe reward hacking because its entropy collapses quickly toward zero. Simultaneously, its response length in the right plot increases uncontrollably, indicating that the model has learned to game the PRM by over-generating verbose and repetitive steps to obtain a higher averaged process reward. As a result, Blend-PRM-GRPO's test accuracy even falls below GRPO. In contrast, PROF maintains a gradual and slightly faster decrease in entropy loss together with controlled response-length growth. This illustrates that our filtration method effectively leverages the PRM signal while staying robust to reward hacking. Below, we further analyze the quality of intermediate reasoning steps.


\subsection{PROF Improves Reasoning Process Quality}

\begin{figure}[t]
    \centering
    \begin{subfigure}[t]{0.48\linewidth}
        \centering
        \includegraphics[width=\linewidth]{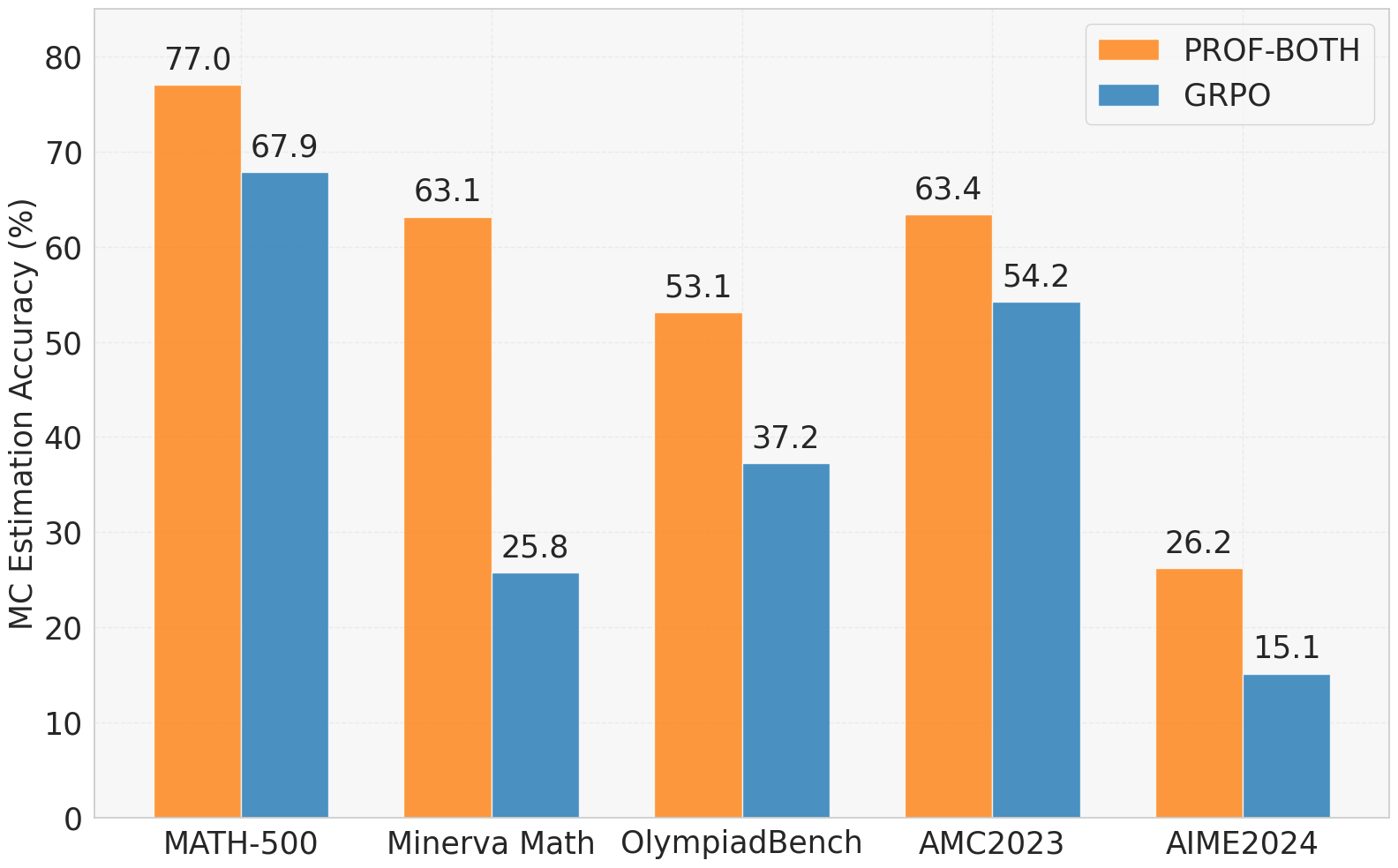}
    \end{subfigure}
    \hfill
    \begin{subfigure}[t]{0.48\linewidth}
        \centering
        \includegraphics[width=\linewidth]{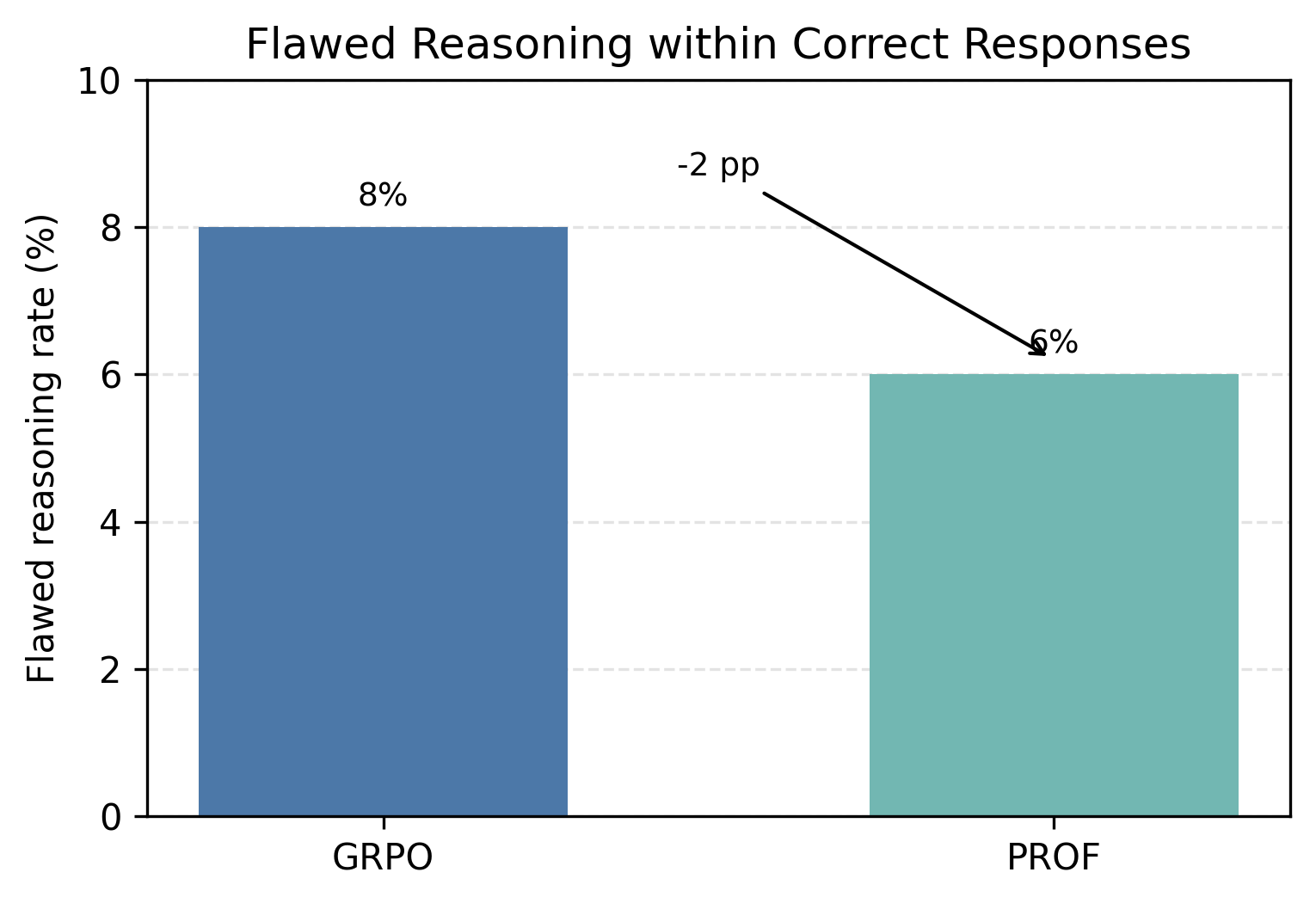}
    \end{subfigure}
    \caption{Reasoning-process quality of PROF vs.\ GRPO. Left: Monte Carlo step-value scores across five benchmarks. Right: flawed-reasoning rate within correct test responses judged by Claude Sonnet 4.6 (GRPO: $8\%$, PROF: $6\%$).}
    \label{fig:understand_prm_shape_reasoning}
\end{figure}

\paragraph{PROF Improves Reasoning Consistency.}
To evaluate the quality of intermediate steps, we adopt Monte Carlo (MC) estimation, a common way to estimate the probability of reaching correct final answers \citep{wang2023math,xiong2024building,luo2024improvemathematicalreasoninglanguage}. For this analysis, we select problem-response pairs from test prompts where our method and GRPO both produced the correct final answer. Both models were initialized from Qwen2.5-Math-7B-base. To estimate the value of each reasoning step, we generate eight independent completions from that point using a temperature of 1.0, and the resulting empirical success rate serves as the MC value. In Figure~\ref{fig:understand_prm_shape_reasoning} (left), the average MC estimates across all five benchmarks are consistently higher for our model. The specific improvement gaps are $9.2\%$ on Math500, $37.4\%$ on Minerva Math, $15.9\%$ on Olympiad Bench, $9.2\%$ on AMC2023, and $11.1\%$ on AIME2024, which are much larger than the outcome-accuracy gap in Table~\ref{tab:main_benchmark}.

\paragraph{PROF Reduces Flawed Reasoning within Correct Responses.}
As a more direct faithfulness metric, we audit correct responses on the test set with Claude Sonnet 4.6 and ask whether the reasoning process contains any flaw (e.g., logical or arithmetic errors), even when the final answer is correct. The audit prompt is provided in Appendix~\ref{s:prompt_template}. In Figure~\ref{fig:understand_prm_shape_reasoning} (right), the flawed-reasoning rate within correct responses decreases from $8\%$ for GRPO to $6\%$ for PROF. This complements Figure~\ref{fig:false_positive_rate}: Figure~\ref{fig:false_positive_rate} measures flawed-reasoning prevalence in base-model outputs before RL (about $30\%$, specifically $26.28\%$), whereas Figure~\ref{fig:understand_prm_shape_reasoning} reports the same notion after training, where both methods fall below $10\%$ and PROF remains lower. We also note that Claude-based auditing is still an approximate signal of reasoning quality and cannot fully replace careful human judgment on step granularity, subtle unsupported jumps, or the level of detail. Therefore, we additionally provide qualitative response comparisons in Figures~\ref{fig:PRM_example_minerval} and \ref{fig:PRM_example_math500}. These examples consistently show that PROF produces concrete and verifiable intermediate deductions, GRPO tends to skip key steps, and Blend-PRM-GRPO is often verbose but less reliable in core calculations.

Additional process metrics on Math500 (step counts and averaged PRM scores) are moved to Appendix Figure~\ref{fig:appendix_process_metrics}. The key takeaway is that PROF improves process quality under both MC-based estimation and direct flaw auditing.

\section{Ablations}
\subsection{Robustness to PRM Capability}
\begin{table*}[h!]
\small
\centering
\setlength{\tabcolsep}{6pt}
\begin{tabular}{c|cccccc}
\hline
Algorithm & Math500 & Minerva Math & Olympiad Bench & AIME24 & AMC23 & Average \\
\hline
GRPO            & 81.6 & 37.2 & 45.5 & 20.6 & 64.4 & 49.9\\
Blend (PRM-7B)        & 81.7 & 36.7 & 45.0 & 15.2 & 58.0 & 47.3 \\
PROF (PRM-7B)      & 83.1 & 39.0 & \textbf{47.8}  & 17.5 & \textbf{70.9} & \textbf{51.7}\\
Blend (PRM-1.5B)   & 81.1 & 37.8 & 44.1 & 11.7 & 62.8 & 47.5\\
PROF-POS (PRM-1.5B)   & 82.9 & \textbf{39.4} & 47.4 & \textbf{19.2} & 66.1 & \underline{51.0}\\
PROF-BOTH (PRM-1.5B) & \textbf{83.2} & 38.8 & \textbf{47.8} & 17.5 & 65.0 & 50.5\\
\hline
\end{tabular}
    \caption{Test accuracy (average@16, temperature $1.0$) for Qwen2.5-Math-7B-base trained with different PRMs, averaged across five benchmarks.}
\label{tab:diff_prm}
\end{table*}

To showcase PROF's robustness to PRM quality, we use a weaker and smaller Skywork-PRM-1.5B \citep{he_2024_16998085} while training from Qwen2.5-Math-7B-base. The results in Table~\ref{tab:diff_prm} validate that when using a weaker PRM, Blend achieves lower accuracies, while PROF still maintains performance close to the model trained with the 7B PRM. This finding further corroborates the robustness of our algorithm.

\subsection{Generality beyond GRPO: RAFT++}
To demonstrate that PROF is a general filtration framework, we extend our experiments to RAFT++ \citep{xiong2025minimalist}, a rejection-sampling-based online training paradigm that only trains on positive samples. We compare PROF-Raft++ against standard RAFT++ baselines with different rollout budgets in Table~\ref{tab:raftpp}. PROF-Raft++ not only outperforms the standard Raft++-$n4$ baseline, but also significantly surpasses Raft++-$n8$. Since RAFT++ only uses positive samples and does not involve negative samples, this comparison is primarily influenced by the number and quality of positive trajectories. Therefore, PROF's priority-based filtration is algorithm-agnostic and consistently identifies high-quality reasoning paths that lead to better policy improvement, regardless of the underlying RL objective.

\begin{table}[t]
\small
\centering
\setlength{\tabcolsep}{8pt}
\begin{tabular}{l|c}
\hline
Method & Average score \\
\hline
Raft++-$n4$ & 35.27 \\
Raft++-$n8$ & 37.75 \\
PROF-Raft++ ($n8m4$) & \textbf{39.29} \\
\hline
\end{tabular}
\caption{RAFT++ ablation under different rollout budgets.}
\label{tab:raftpp}
\end{table}

\subsection{Separating Correct and Incorrect}
\begin{figure}[t]
    \centering
    \begin{subfigure}[t]{0.48\linewidth}
        \centering
        \includegraphics[width=\linewidth]{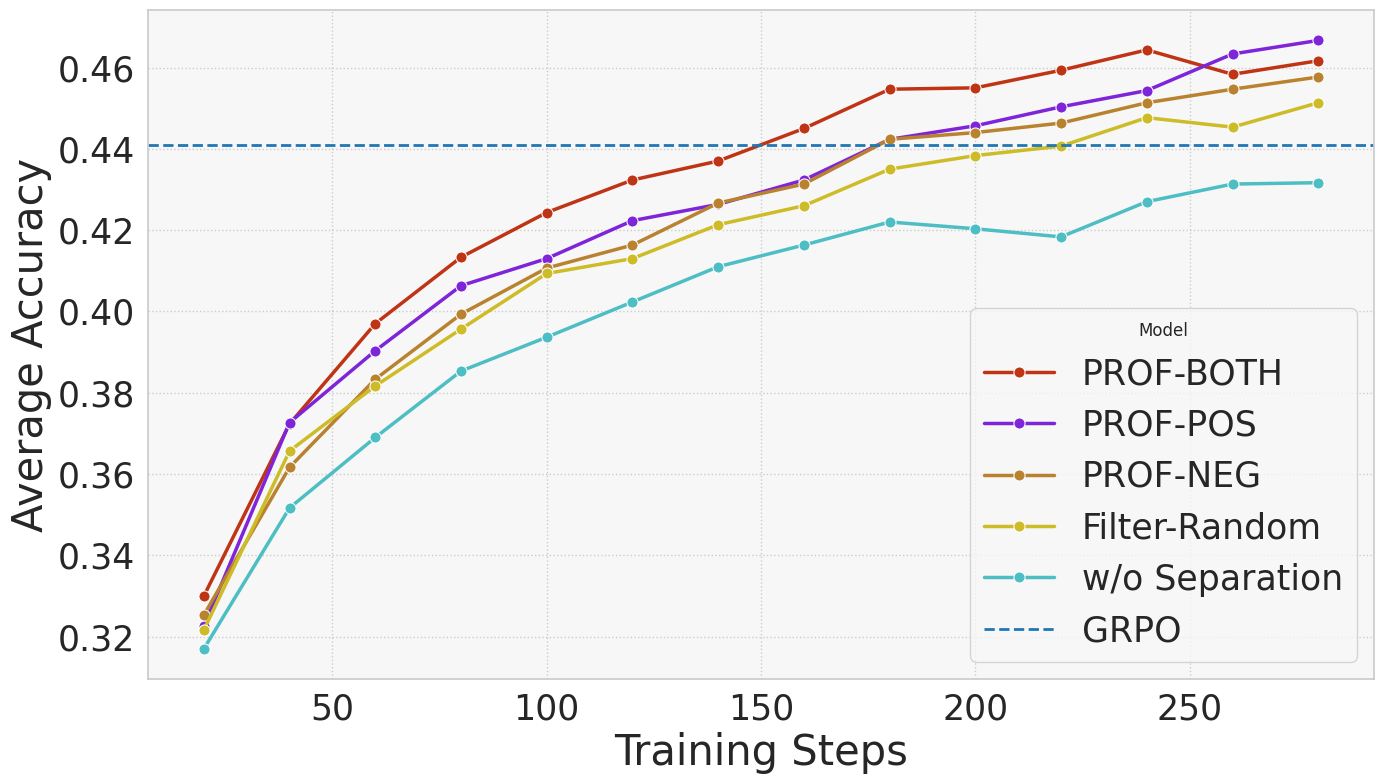}
    \end{subfigure}
    \hfill
    \begin{subfigure}[t]{0.48\linewidth}
        \centering
        \includegraphics[width=\linewidth]{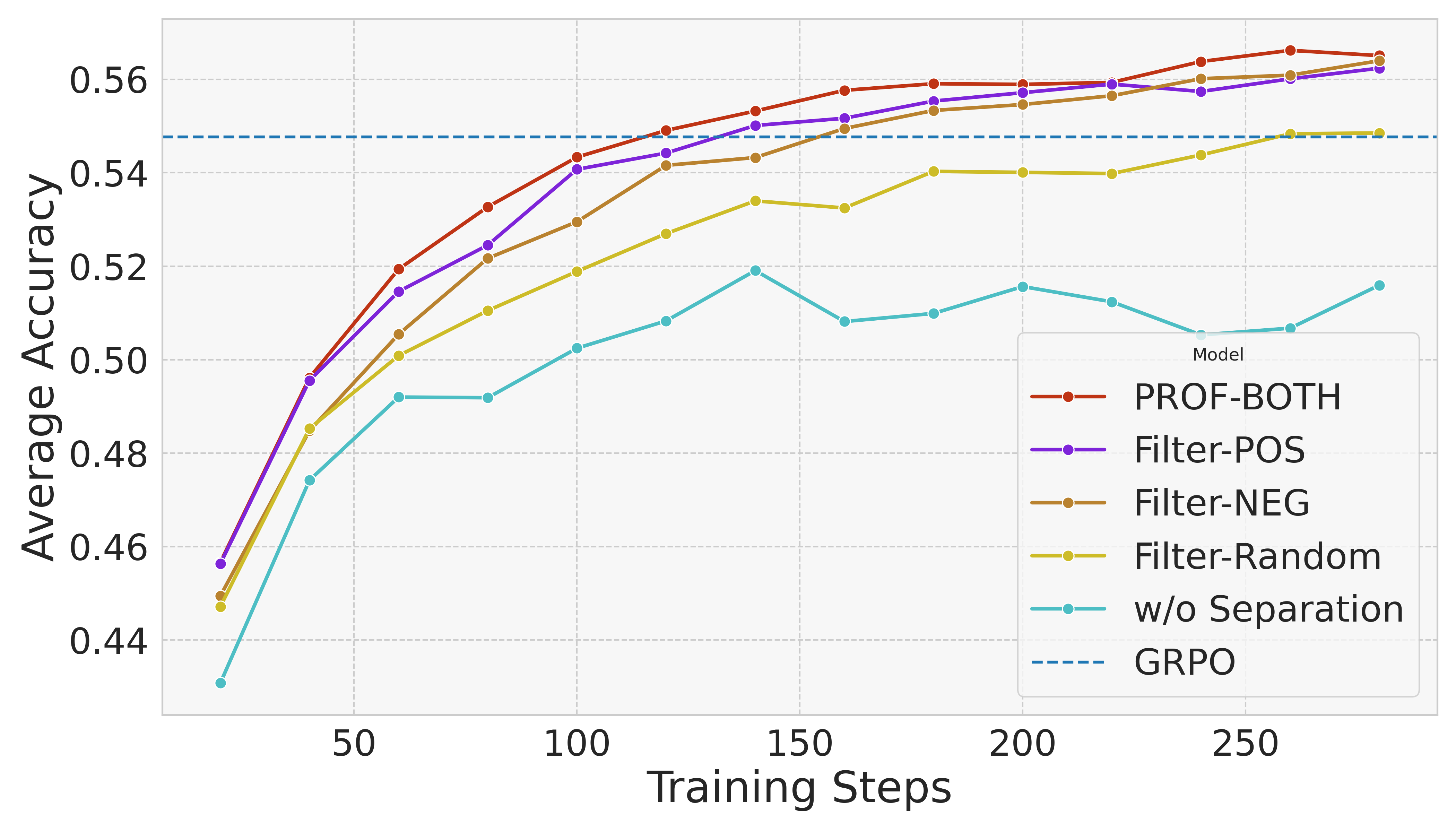}
    \end{subfigure}

    \begin{subfigure}[t]{0.62\linewidth}
        \centering
        \includegraphics[width=\linewidth]{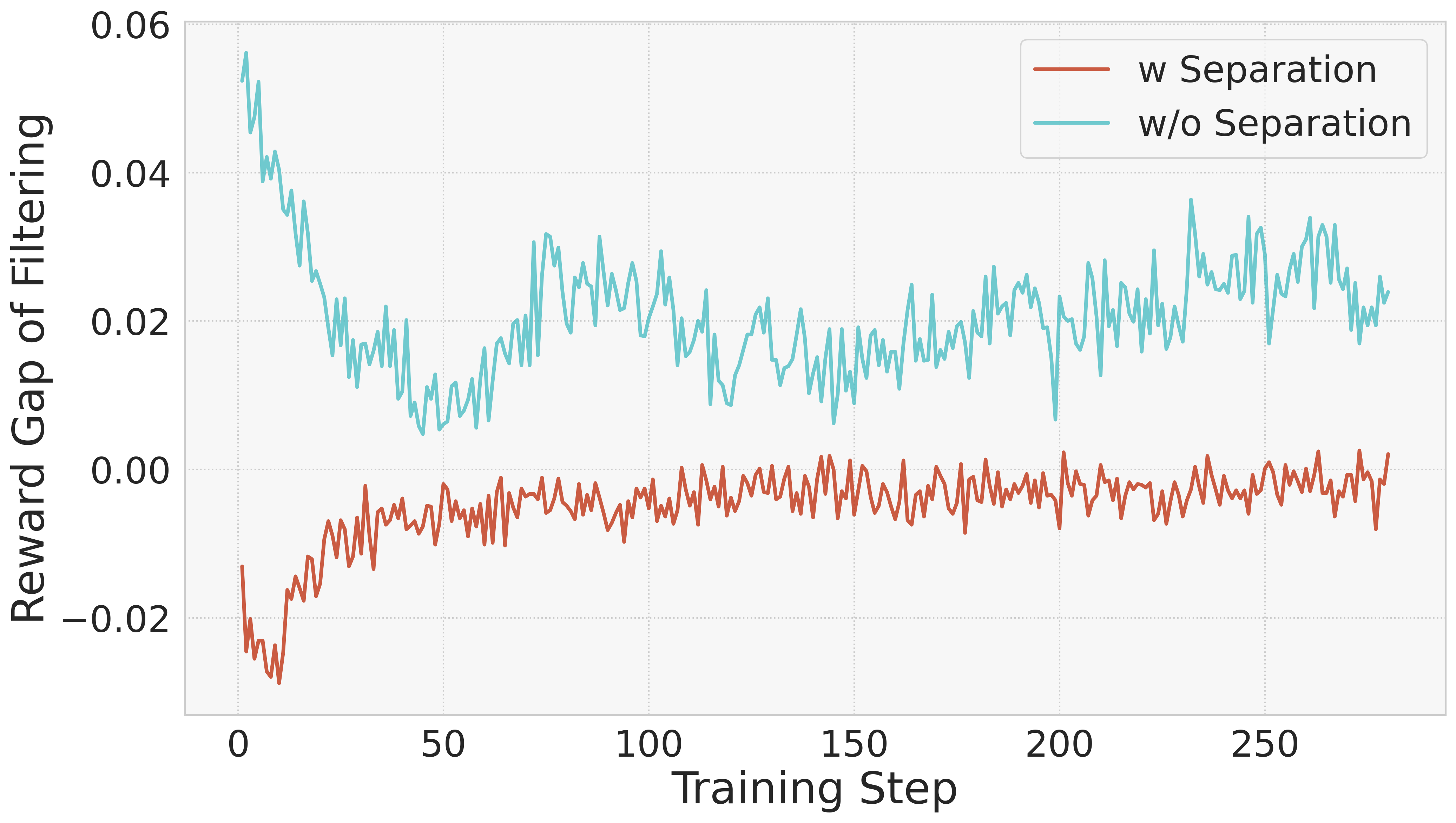}
    \end{subfigure}
    \caption{Left two: averaged accuracy on Math500, Minerva Math, and Olympiad Bench for PROF variants from Qwen2.5-Math-1.5B-base and Qwen2.5-Math-7B-base. Right: reward gap after vs.\ before filtering, with and without separation.}
    \label{fig:correct_incorrect}
\end{figure}

We first test a no-separation variant (PROF w/o separation) that ranks all rollouts together. To mitigate PRM scale bias, we center each step score by subtracting the batch mean. Even with centering, the rightmost plot in Figure~\ref{fig:correct_incorrect} shows that PROF w/o separation has over $2\%$ gap between rewards before and after filtering, indicating disproportionate removal of negative samples. A likely reason is that incorrect responses often contain several locally correct steps, which can inflate averaged PRM scores and blur process-outcome consistency. Separating correct and incorrect groups alleviates this bias.

We then compare three variants: PROF-POS (consistency filtering on correct group only), PROF-NEG (incorrect group only), and Filter-Random (random filtering on both groups) \citep{xu2025not}. As shown in Figure~\ref{fig:correct_incorrect}, PROF-POS and PROF-BOTH are the best-performing strategies across both 1.5B and 7B settings; PROF-BOTH is typically more sample-efficient, PROF-NEG is weaker, Filter-Random is only slightly above GRPO, and w/o separation is the worst. These results suggest that preserving quality in correct responses is the dominant factor, while consistency control on incorrect responses is secondary. More filtration ablations are provided in Appendix~\ref{s:additional_experimental_results}.

This ablation highlights a practical trade-off between PROF-BOTH and PROF-POS. PROF-BOTH usually converges faster by using consistency signals from both groups, while PROF-POS can be more robust when PRM reliability is weaker or distribution shift is larger, since it avoids tightly shaping the incorrect group with noisy estimates. In both cases, improving correct trajectories is the main driver, and filtering incorrect trajectories mainly affects efficiency and stability.

\section{Conclusion and Future Work}
This work introduces Process Consistency Filter (PROF), a data curation technique that filters generated responses based on PRM--ORM consistency while maintaining a balanced correct/incorrect ratio. We demonstrate that PROF consistently improves final-answer accuracy and shapes the policy to generate more detailed and fine-grained intermediate reasoning steps. PROF is also a general filtration framework rather than one tied to a specific PRM or RL objective. Thus, using pre-trained PRMs in our experiments is not a limitation; instead, it highlights the robustness of our algorithm to different PRMs and suggests that training a task-specific PRM for each base model is unnecessary. Exploring stronger or more diverse PRMs, and extending PROF to other reasoning tasks such as coding \citep{jimenez2023swe} and web navigation \citep{zhou2023webarena}, remains important future work.

\paragraph{Broader Impact and Ethics Statement} Our work contributes to AI safety by enhancing the faithfulness and interpretability of chain-of-thought reasoning, mitigating the risk of misleading hallucinations. However, we acknowledge two potential risks. First, the reliance on oversampling and dense process reward computation increases computational overhead and environmental impact compared to standard baselines. Second, our filtration mechanism depends on pre-trained Process Reward Models (PRMs); if these PRMs harbor biases toward specific reasoning patterns or languages, our method may inadvertently amplify such biases by filtering out diverse but valid solutions. We encourage future research to address these efficiency and fairness challenges.

\section*{Limitations}
Although PROF can effectively improve robustness to PRM noise and increase reasoning-step quality, our method requires more computation than Blend or vanilla GRPO because it first oversamples and then filters. How to balance efficiency and reasoning quality remains an important direction for future work. Finally, we acknowledge the use of AI assistants (e.g., ChatGPT) for grammatical error correction and polishing of the manuscript.



\bibliography{custom}

\newpage

\appendix

\section{Additional Experimental Details and Results}\label{sec:additional_experimental_details}

\subsection{Main Experiments}
The implementations are based on the verl framework \citep{sheng2025hybridflow}, and we follow most of its parameter settings. Specifically, we use the AdamW optimizer with learning rate $1\times 10^{-6}$. We adopt the clip-higher trick \citep{yu2025dapo}, which clips the sampling ratio $\pi_\theta/\pi_{\text{old}}$ to an asymmetric range $(1-\epsilon_{\text{low}}, 1+\epsilon_{\text{high}})$. Specifically, we set $\epsilon_{\text{low}}=0.2,\epsilon_{\text{high}}=0.28$ for models initialized from Qwen2.5-Math-1.5B-base and maintain $\epsilon_{\text{high}}=\epsilon_{\text{low}}=0.2$ for other cases. In each iteration, we sample $1024$ prompts and roll out $n=4$ responses per prompt for GRPO and $n=8$ responses for PROF. Note that the policy update number for all algorithms is $m=4$. For the regularization of step numbers in Algorithm~\ref{alg}, we take $\lambda=10$ and $H_\lambda=30$.
For the rollout stage, we use a temperature of $1.0$ and a top-p value of $1.0$. We set the KL loss coefficient to $0.001$ and entropy loss coefficient to $0.001$. All the models are trained with $8$ H100 GPUs. We set the training mini-batch size as $256$ and allow the models to generate $4096$ tokens per prompt.

\begin{figure*}[t]
    \centering
    \begin{subfigure}[t]{0.45\textwidth}
        \centering
        \includegraphics[width=\textwidth]{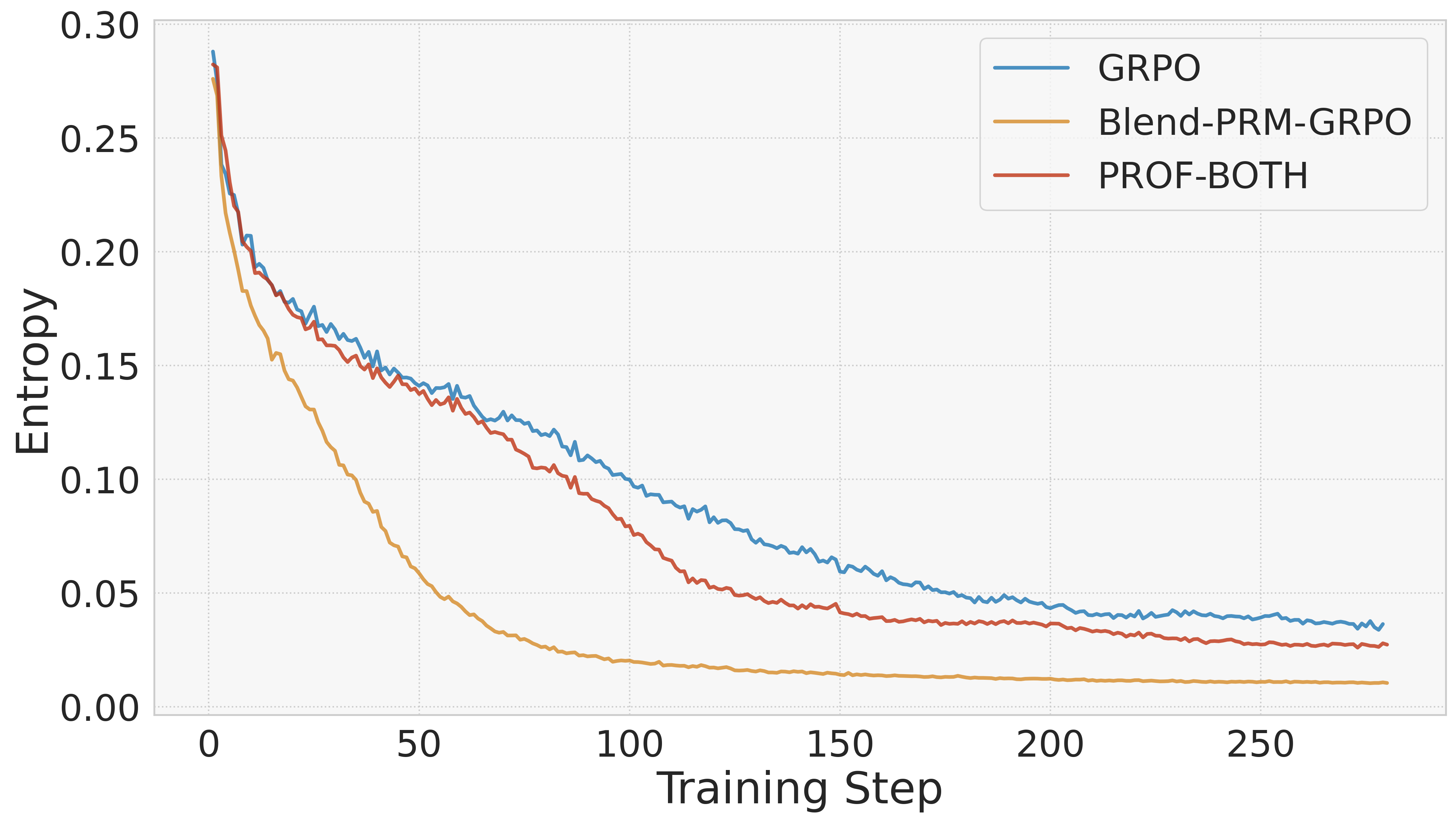}
    \end{subfigure}
    \hfill
    \begin{subfigure}[t]{0.45\textwidth}
        \centering
        \includegraphics[width=\textwidth]{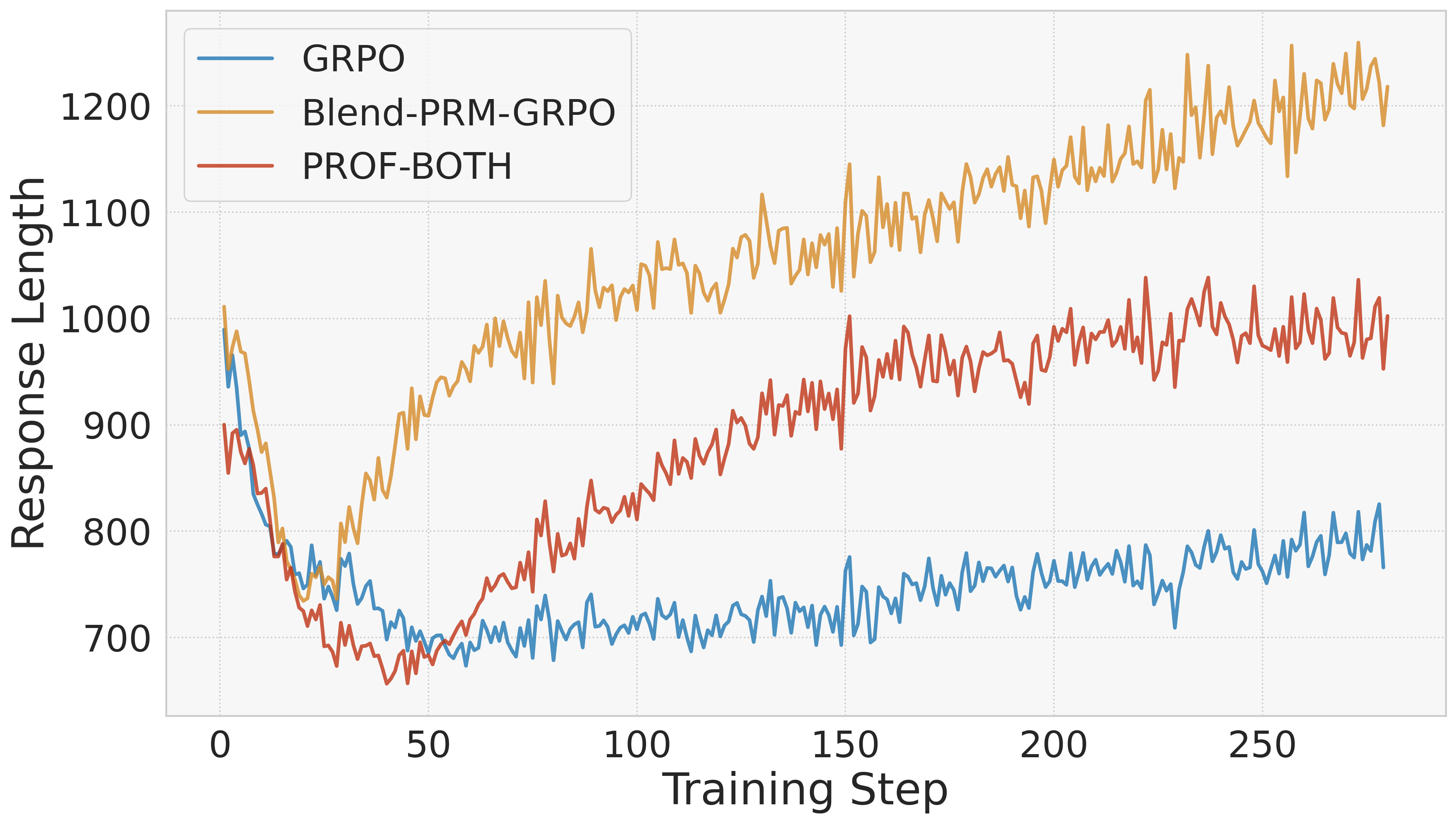}
    \end{subfigure}
    \caption{Entropy loss (left) and response length (right) of the models initialized from Qwen2.5-Math-7B-base.}
    \label{fig:ent}
\end{figure*}

\subsection{Prompt Template}\label{s:prompt_template}
We present the template used by the LLM to audit whether a correct response still contains reasoning flaws in Table~\ref{tab:llm_judge_preference_feedback}.

\begin{table*}[t]
\begin{tcolorbox}[
  colframe=blue!75!black,
  colback=blue!5,
  colbacktitle=blue!20,
  coltitle=black,
  fonttitle=\bfseries,
  boxrule=1pt,
  arc=5pt,
  title=Prompt for Auditing Reasoning Flaws in a Candidate Solution
]
\textbf{System} Your task is to audit a candidate solution to a math problem and determine whether the reasoning process contains any flaw or error, even if the final answer is correct.\\
\textbf{User} You must evaluate the solution step by step and focus on the validity of the reasoning process, not writing style. A solution can be flawed even when the final answer is correct.\\
\textbf{Reasoning flaws include:}
\begin{itemize}
    \item random guessing;
    \item skipping or jumping over intermediate steps;
    \item logical errors;
    \item arithmetic errors;
    \item misuse of definitions, theorems, or formulas.
\end{itemize}
\textbf{Do not penalize:} minor wording issues or overly specific solution styles.\\
\textbf{Important rules:}
\begin{itemize}
    \item Judge the reasoning process itself, not just the final answer.
    \item If the final answer is correct but any earlier step is invalid or unsupported, mark the reasoning as flawed.
    \item Identify the earliest step where the reasoning first becomes flawed.
    \item Once an earliest flawed step is found, later steps may be unreliable because they can depend on that error.
\end{itemize}
You will be given a math problem and a candidate solution. If the candidate solution does not explicitly label steps, first segment it into natural reasoning steps before judging.\\
Return your judgment in exactly the following format:\\
\texttt{<final\_answer\_correct>yes / no / unclear</final\_answer\_correct>}\\
\texttt{<reasoning\_flawed>yes / no</reasoning\_flawed>}\\
\texttt{<earliest\_flawed\_step>step number or none</earliest\_flawed\_step>}\\
\texttt{<error\_type>arithmetic | algebra | logical | unsupported\_jump | random\_guessing | theorem\_misuse | missing\_case | contradiction | answer\_first\_rationalization | none</error\_type>}\\
\texttt{<brief\_explanation>}\\
A concise explanation of why the reasoning is valid, or why the earliest flawed step is flawed.\\
\texttt{</brief\_explanation>}\\
Now evaluate the following example.\\
\texttt{[Problem]}\\
\texttt{\{\{prompt\}\}}\\
\texttt{[Candidate Solution]}\\
\texttt{\{\{responses\}\}}
\end{tcolorbox}
    \caption{Prompt for auditing flawed reasoning in candidate solutions via LLM-as-a-judge.}
    \label{tab:llm_judge_preference_feedback}
\end{table*}

\section{Additional Experimental Results}\label{s:additional_experimental_results}
In this section, we include additional ablation studies and evaluation results for a more comprehensive understanding of the PROF framework.

\subsection{Matched-Cost Results for Difficulty Levels 1--3}
\begin{figure*}[t]
    \centering
    \begin{subfigure}[t]{0.32\textwidth}
        \centering
        \includegraphics[width=\textwidth]{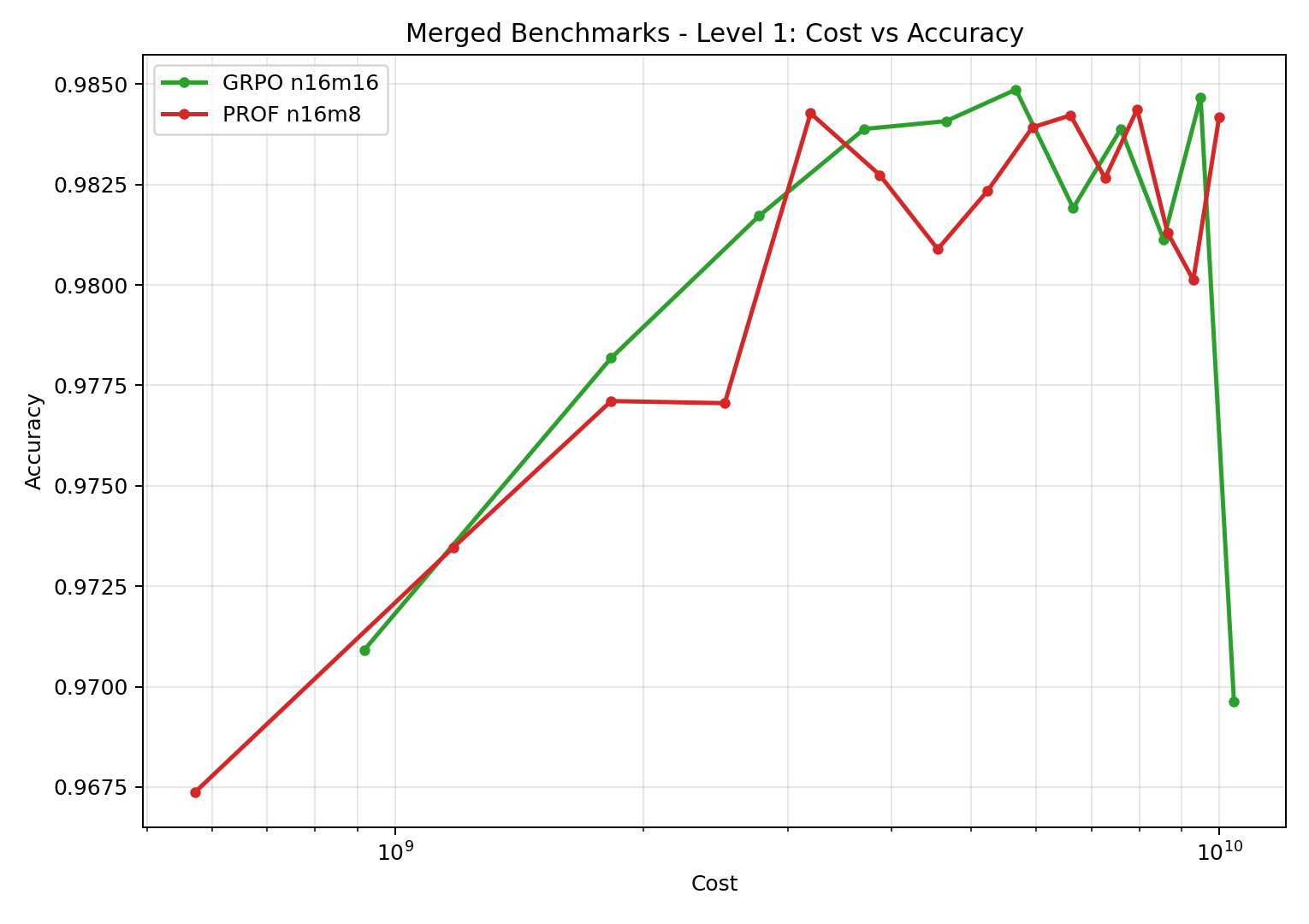}
    \end{subfigure}
    \hfill
    \begin{subfigure}[t]{0.32\textwidth}
        \centering
        \includegraphics[width=\textwidth]{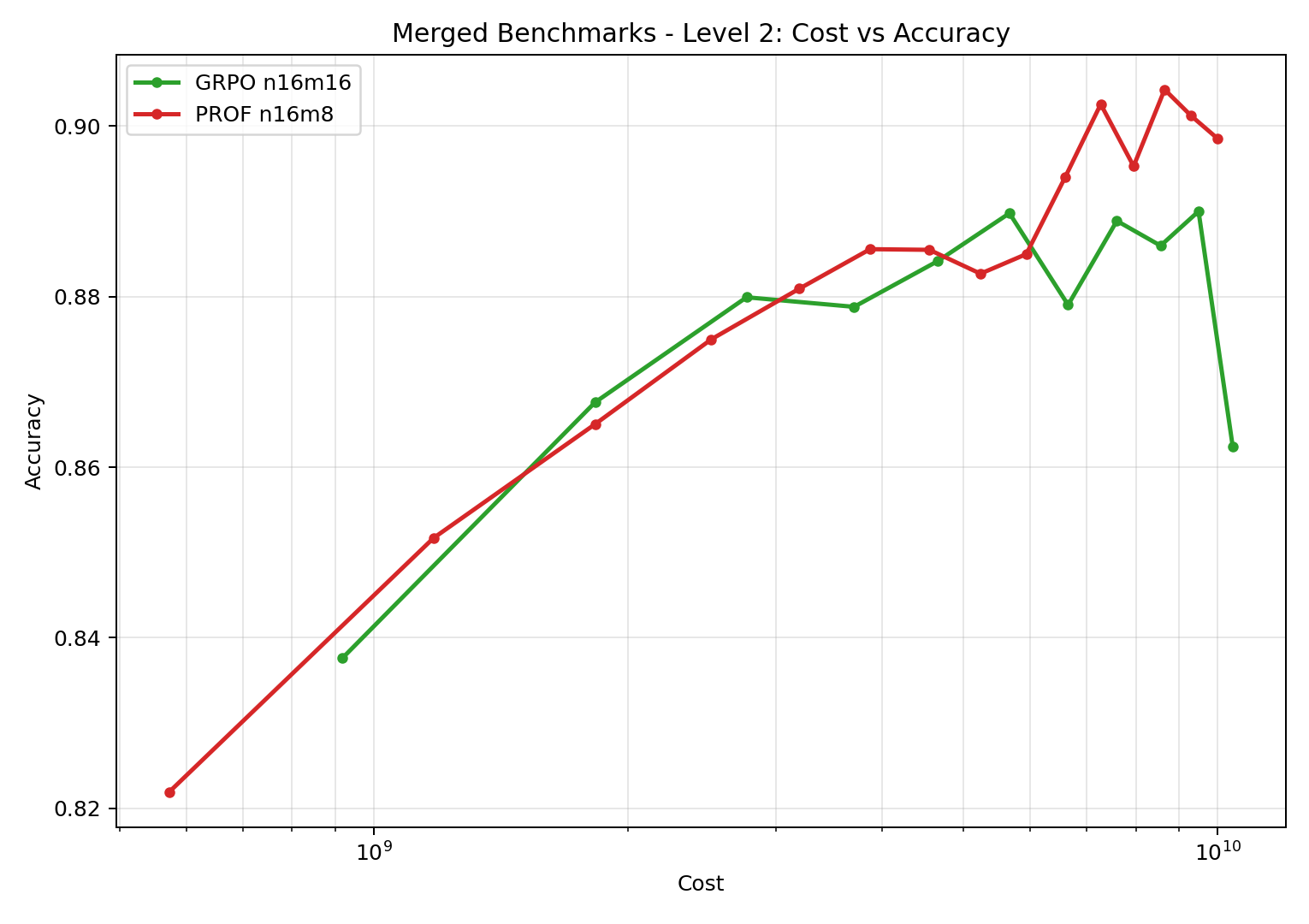}
    \end{subfigure}
    \hfill
    \begin{subfigure}[t]{0.32\textwidth}
        \centering
        \includegraphics[width=\textwidth]{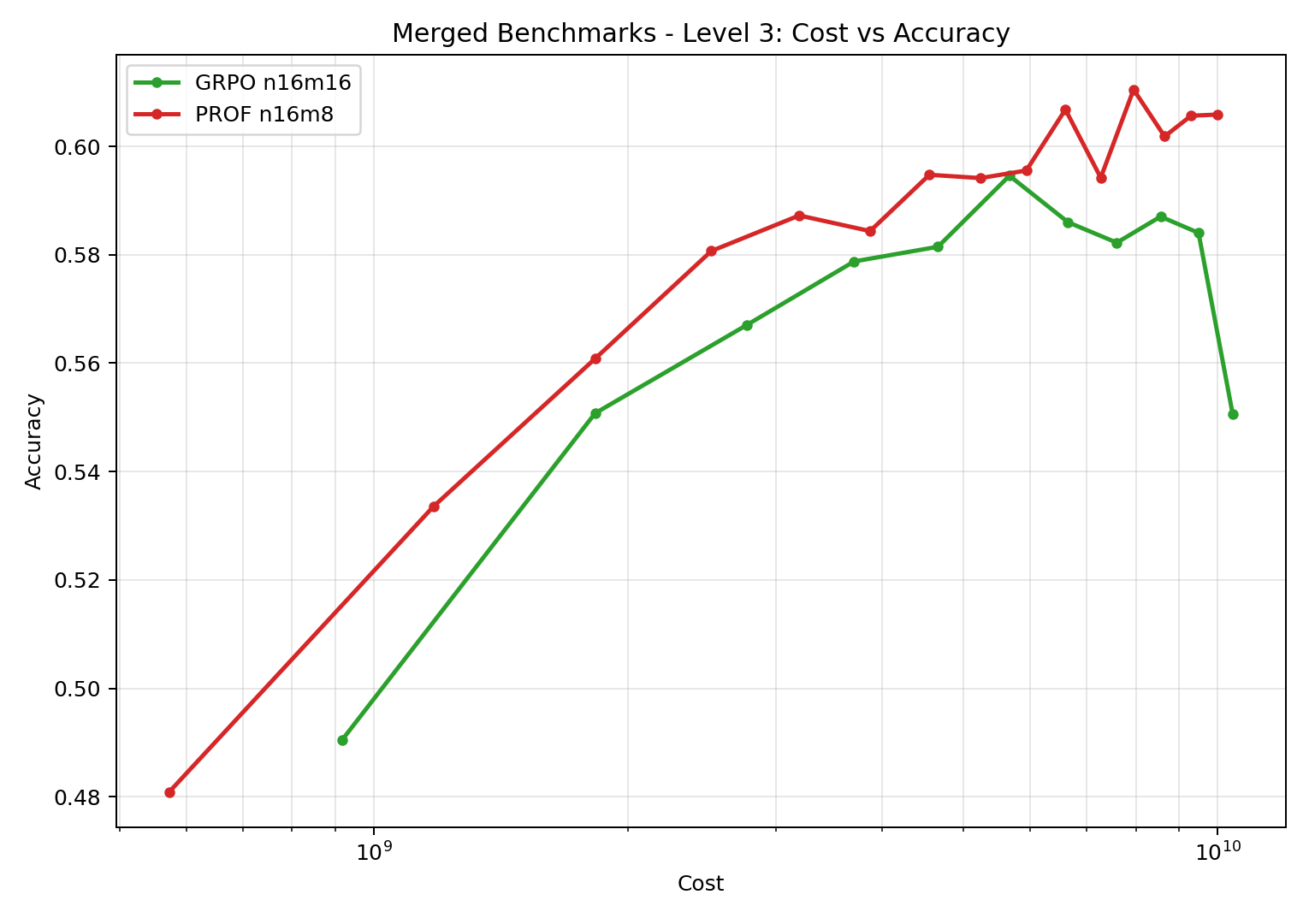}
    \end{subfigure}
    \caption{Matched-cost comparison on Qwen2.5-Math-7B-base for difficulty Level 1 ($p>0.5$), Level 2 ($0.25<p\le0.5$), and Level 3 ($0<p\le0.25$). Average cost is computed as Inference $+ 3\times$ Train $+$ PRM.}
    \label{fig:cost_level_123_appendix}
\end{figure*}

\subsection{Effect of Rollout Numbers}
    \begin{figure}[h!]
        \centering
        \includegraphics[width=0.5\textwidth]{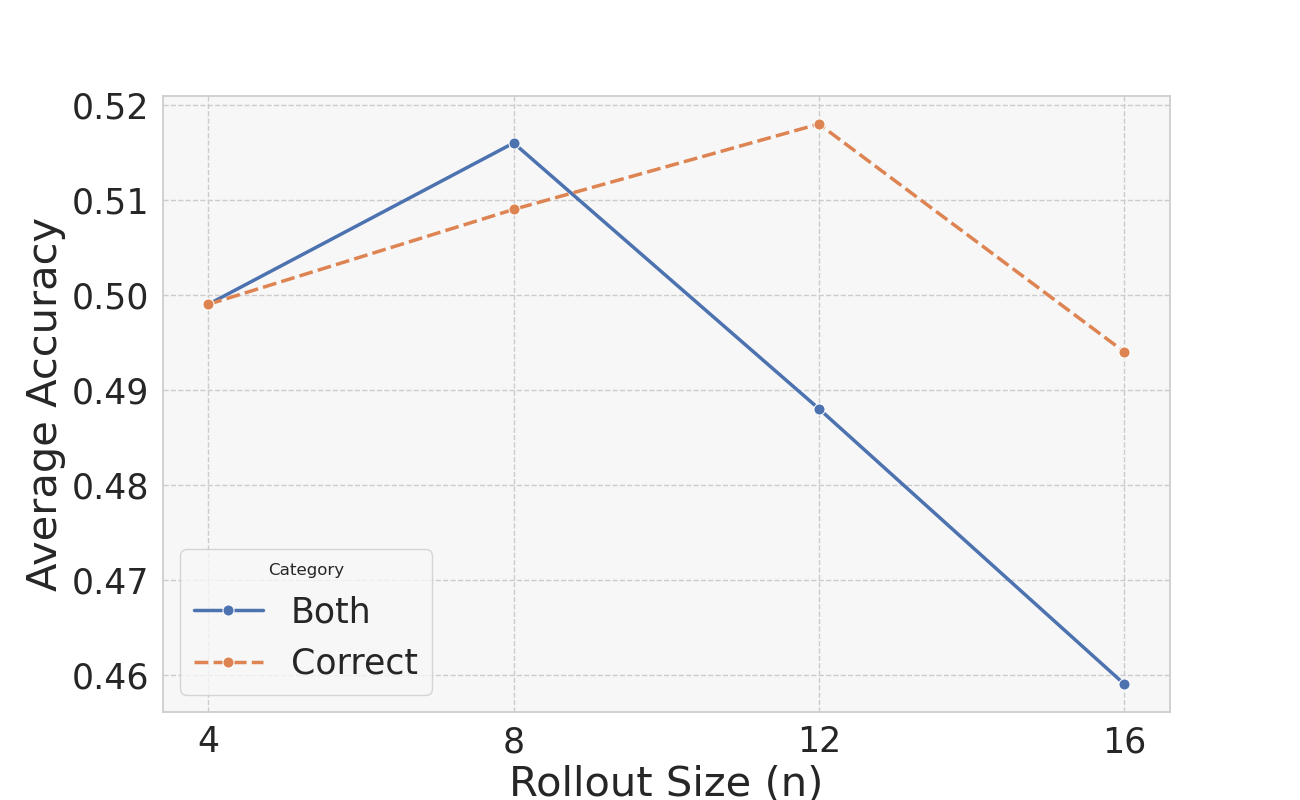}
        \caption{The averaged accuracy across all five benchmarks over rollout sizes $n=4,8,12,16$ for filtering both correct and incorrect groups with PRM consistency (Both) and only the correct group with PRM consistency (Correct).}
        \label{fig:nroll_plot}
    \end{figure}

We study the scale of rollout numbers $n$ with fixed policy-update number $m=4$ by varying $n=4,8,12,16$. The lower-right plot in Figure~\ref{fig:nroll_plot} presents the test accuracy averaged over all five benchmarks for PROF-BOTH (Both) and PROF-POS (Correct) initialized from Qwen2.5-Math-7B-base. We observe that performance first increases and then decreases as $n$ grows, revealing a trade-off between enhancing process reasoning quality and avoiding reward hacking. Notably, PROF-POS decreases later (after $n=12$) because it only leverages PRM influence in the correct group, indicating that PROF-POS is more robust when PRM influence becomes stronger, such as when the ranking-and-filtering scale increases.

\subsection{Additional Process Metrics on Math500}
\begin{figure}[t]
    \centering
    \begin{subfigure}[t]{0.48\linewidth}
        \centering
        \includegraphics[width=\linewidth]{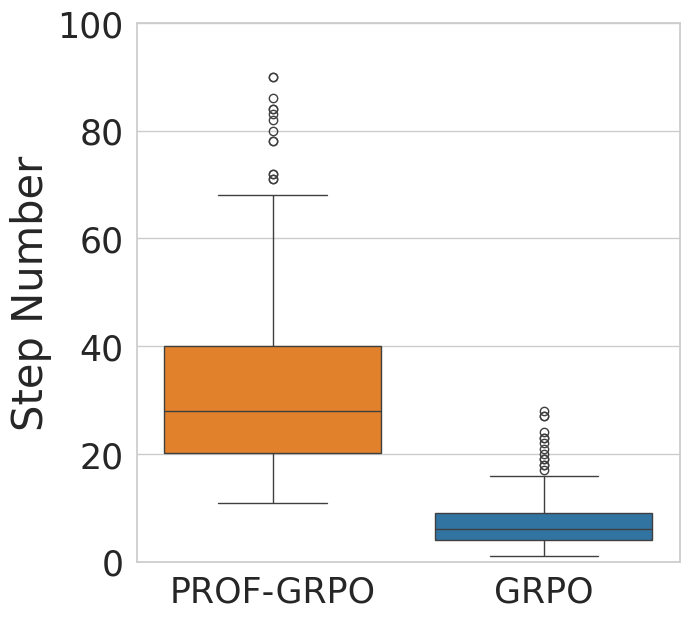}
    \end{subfigure}
    \hfill
    \begin{subfigure}[t]{0.48\linewidth}
        \centering
        \includegraphics[width=\linewidth]{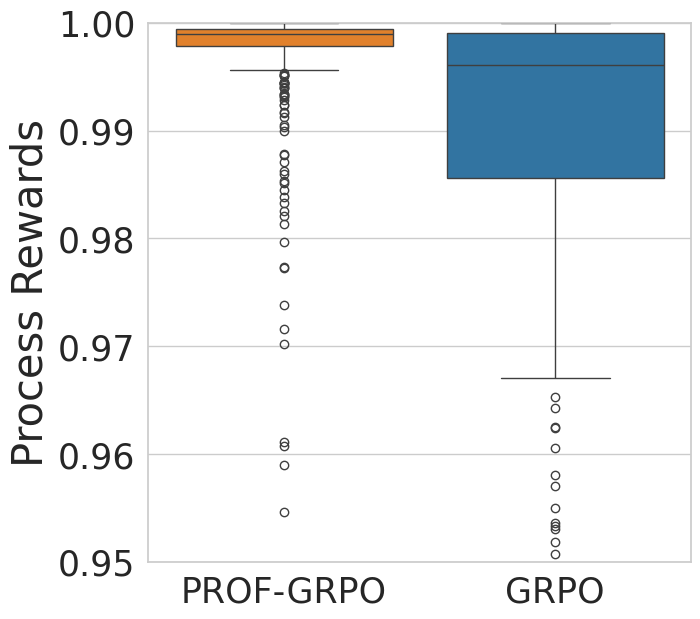}
    \end{subfigure}
    \caption{Supplementary process metrics on Math500 for PROF vs.\ GRPO: number of reasoning steps (left) and averaged Qwen2.5-Math-PRM-7B scores (right).}
    \label{fig:appendix_process_metrics}
\end{figure}

\subsection{Variants of Filtration Methods}

\begin{table*}[t]
\centering
\setlength{\tabcolsep}{6pt}
\begin{tabular}{c|cccccc}
\hline
Algorithm & Math500 & Minerva Math & Olympiad Bench & AIME24 & AMC23 & Average \\
\hline
Mean & \textbf{83.1} & \textbf{39.0} & \textbf{47.8} & 17.5 & \textbf{70.9} & \textbf{51.7} \\
Minimum & 82.9 & 38.3 & 46.7 & 20.8 & 65.9 & 50.9\\
Sum & 82.4 & 38.1 & 47.4 & 17.7 & 67.5 & 50.6\\
Ratio & 81.4 & 36.6 & 45.0 & \textbf{24.8} & 65.2 & 50.6\\
\hline
\end{tabular}
\caption{Performance of different filtration ways in PROF starting from Qwen2.5-Math-7B-base.}
\label{tab:diff_compute_consistency}
\end{table*}

In this subsection, we investigate different ways of computing the consistency score $r^{\mathrm{pro}}$, in addition to taking the mean of PRM scores over steps. Here, Mean denotes averaging over steps in Algorithm~\ref{alg}; Minimum and Sum denote taking the minimum and the sum over steps; and Ratio denotes filtering while preserving the original positive/negative sample distribution instead of balancing it. As shown in Table~\ref{tab:diff_compute_consistency}, Minimum ($50.9\%$), Sum ($50.6\%$), and Ratio ($50.6\%$) all underperform Mean. This suggests that the mean provides a more stable estimate of reasoning consistency: unlike the minimum, it is less sensitive to a single poorly scored step, and unlike the sum, it avoids bias toward longer trajectories. Additionally, balancing the correct/incorrect ratio lets consistency-based filtering select the better-supported group without breaking class balance.

\subsection{Effect of Step Number}
To verify that PROF does not help merely by increasing the number of steps, we evaluate Filter-Nstep, which ranks and filters samples by shorter step counts instead of lower PRM--ORM consistency.

\begin{figure}[h!]
    \centering
    \includegraphics[width=1.0\linewidth]{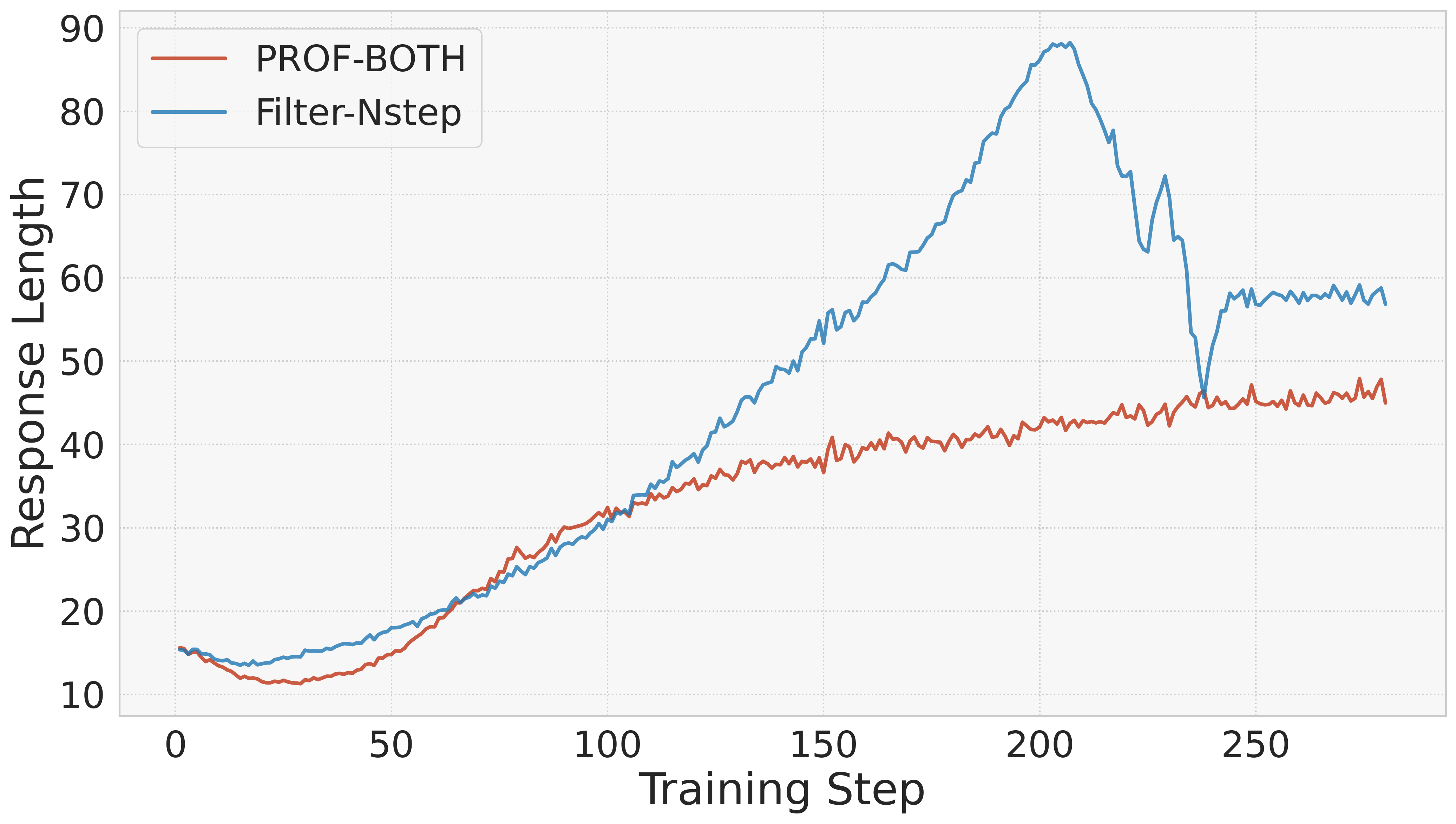}
    \caption{The number of reasoning steps during training time for PROF-GRPO and Filter-Nstep initialized from Qwen2.5-Math-7B-base.}
    \label{fig:len_segs_mean}
\end{figure}

From Table~\ref{tab:diff_compute_consistency}, we find that Ratio scores only $50.6\%$ on average and cannot compete with balanced filtering (PROF), which further corroborates the importance of maintaining a balanced correct/incorrect proportion. Additionally, because PROF increases the number of intermediate reasoning steps, we compare against simple step-length filtering to verify that the gain does not come merely from longer responses. As shown in Figure~\ref{fig:len_segs_mean} and Table~\ref{tab:variants}, Filter-Nstep mainly manipulates step length, exhibits an unreasonable increase followed by a sudden drop, and yields inferior average accuracy.

\begin{table*}[t]
\centering
\setlength{\tabcolsep}{6pt}
\begin{tabular}{c|cccccc}
\hline
Algorithm & Math500 & Minerva Math & Olympiad Bench & AIME24 & AMC23 & Average \\
\hline
PROF-BOTH & \textbf{83.1} & \textbf{39.0} & \textbf{47.8} & 17.5 & \textbf{70.9} & \textbf{51.7} \\
Filter-Nstep & 81.5 & 35.5 & 45.9 & 16.3 & 58.6 & 47.6\\
\hline
\end{tabular}
\caption{Performance of filtration variants besides PROF-BOTH for Qwen2.5-Math-7B-base, averaged over all five benchmarks. Ratio preserves the original correct/incorrect proportion, and Filter-Nstep ranks and filters by the number of step segments.}
\label{tab:variants}
\end{table*}

\section{Additional Examples}

\begin{figure*}[htp]
    \centering
    \includegraphics[width=0.8\textwidth]{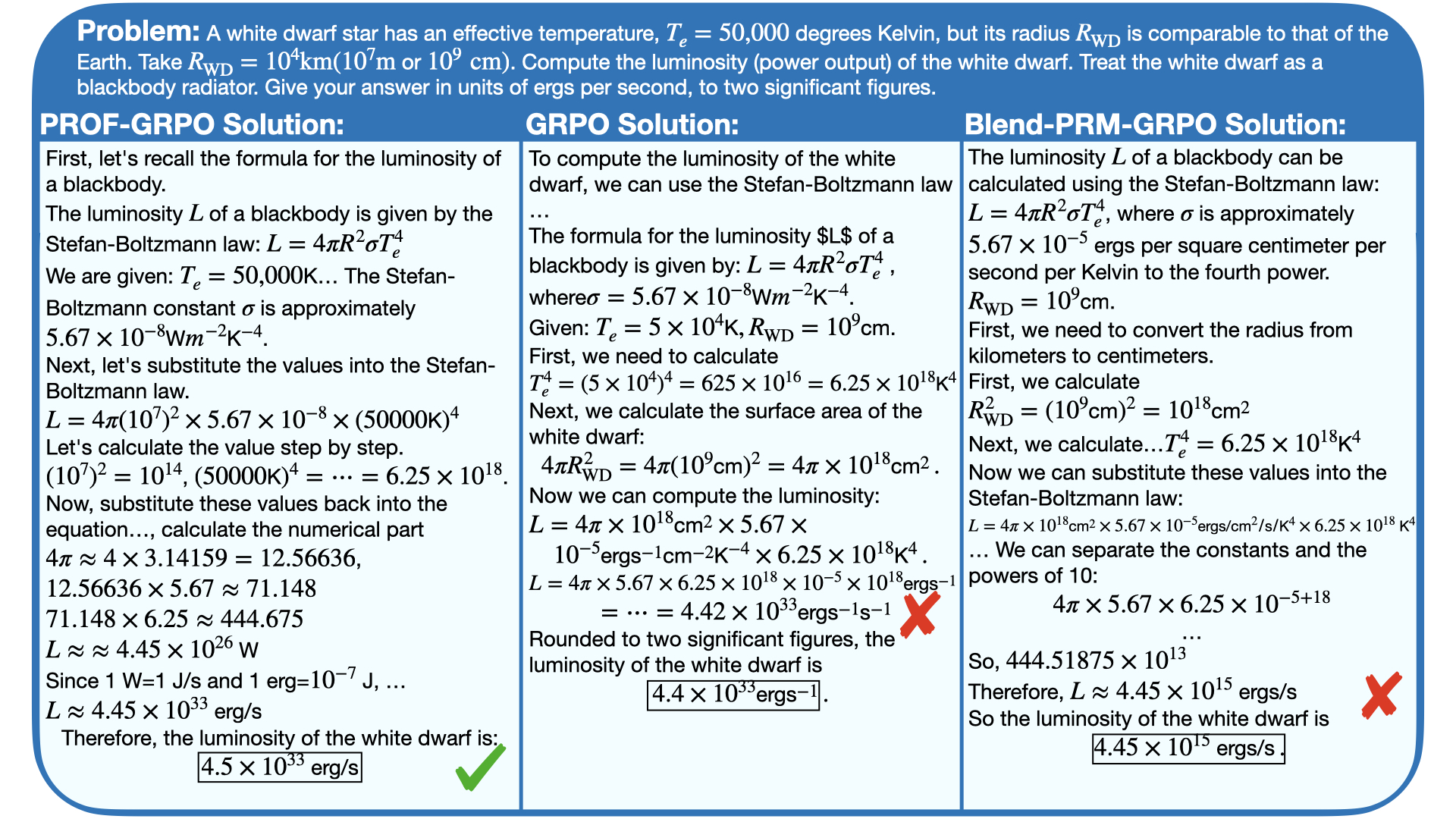}
    \caption{A Minerva-Math example comparing distinct intermediate reasoning patterns of PROF-BOTH, vanilla GRPO, and Blend-PRM-GRPO. PROF-BOTH presents concrete and correct deduction steps. GRPO's solution skips detailed deductions and contains flaws in calculation precision and final rounding. Blend-PRM-GRPO is long-winded and makes a major mistake in computing the power.}
    \label{fig:PRM_example_minerval}
\end{figure*}

\begin{figure*}[h]
    \centering
    \includegraphics[width=0.8\textwidth]{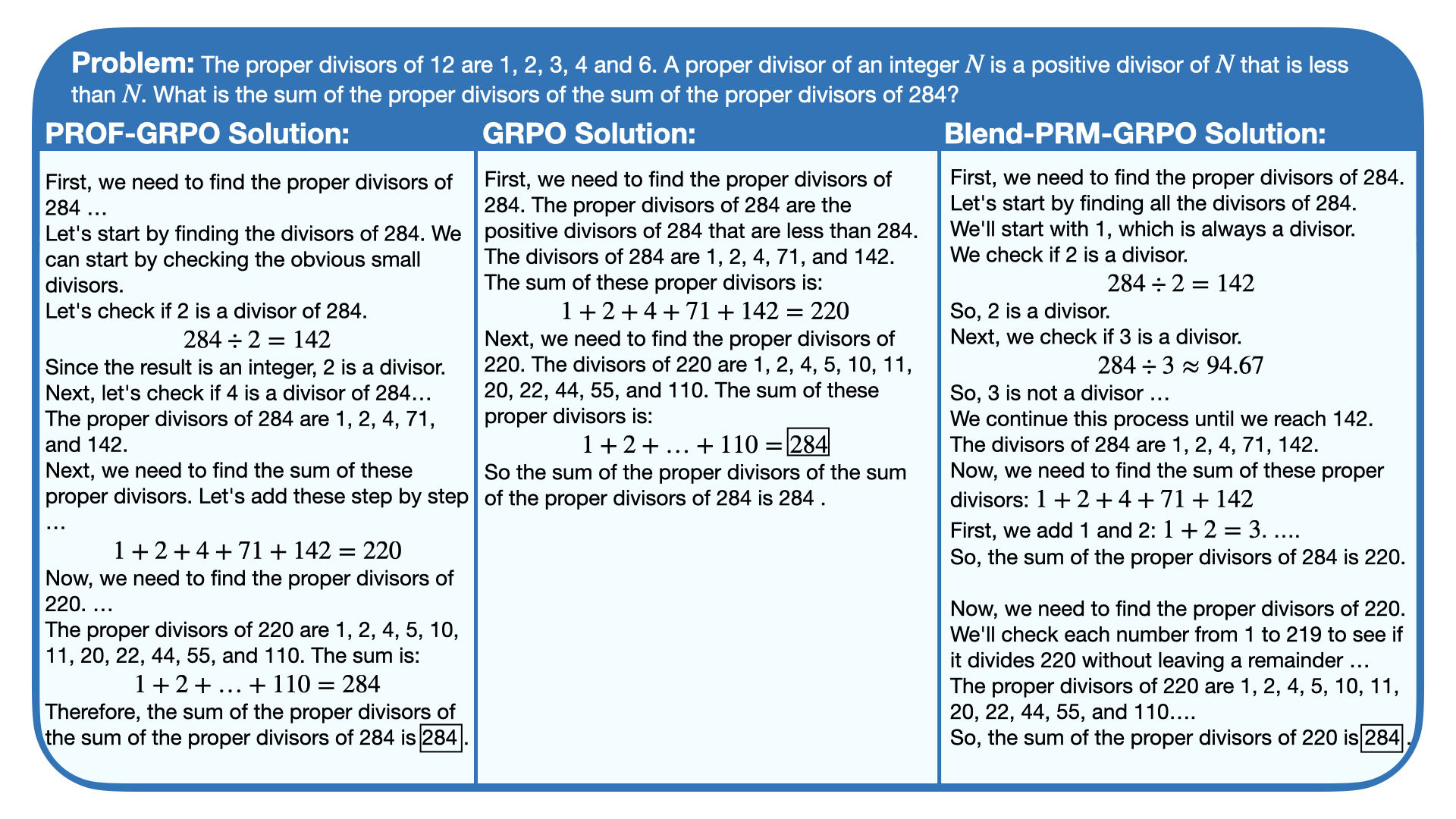}
    \caption{A Math500 example to compare distinct intermediate reasoning patterns of PROF-BOTH, vanilla GRPO and Blend-PRM-GRPO. PROF-BOTH presents concrete and correct deduction steps. PROF-BOTH's solution shows how to find the divisors and summation in detail, and is easy to follow. GRPO skips all core reasoning. Blend-PRM-GRPO has inefficient and excessively tedious steps.}
    \label{fig:PRM_example_math500}
\end{figure*}

\end{document}

%% file: math_commands.tex
\usepackage{amsmath,amsfonts,bm}

\def\eqref#1{equation~\ref{#1}}

\def\1{\bm{1}}

\DeclareMathAlphabet{\mathsfit}{\encodingdefault}{\sfdefault}{m}{sl}
\SetMathAlphabet{\mathsfit}{bold}{\encodingdefault}{\sfdefault}{bx}{n}



%% file: custom.bib
@misc{he_2024_16998085,
  author       = {He, Jujie and
                  Wei, Tianwen and
                  Yan, Rui and
                  Liu, Jiacai and
                  Wang, Chaojie and
                  Gan, Yimeng and
                  Tu, Shiwen and
                  Liu, Chris Yuhao and
                  Zeng, Liang and
                  Wang, Xiaokun and
                  Wang, Boyang and
                  Li, Yongcong and
                  Zhang, Fuxiang and
                  Xu, Jiacheng and
                  An, Bo and
                  Liu, Yang and
                  Zhou, Yahui},
  title        = {Skywork-o1 Open Series},
  month        = nov,
  year         = 2024,
  publisher    = {Zenodo},
  version      = {1.0.0},
  doi          = {10.5281/zenodo.16998085},
  url          = {https://doi.org/10.5281/zenodo.16998085},
}

@article{baker2025monitoring,
  title={Monitoring reasoning models for misbehavior and the risks of promoting obfuscation},
  author={Baker, Bowen and Huizinga, Joost and Gao, Leo and Dou, Zehao and Guan, Melody Y and Madry, Aleksander and Zaremba, Wojciech and Pachocki, Jakub and Farhi, David},
  journal={arXiv preprint arXiv:2503.11926},
  year={2025}
}

@article{chen2025reasoning,
  title={Reasoning Models Don't Always Say What They Think},
  author={Chen, Yanda and Benton, Joe and Radhakrishnan, Ansh and Uesato, Jonathan and Denison, Carson and Schulman, John and Somani, Arushi and Hase, Peter and Wagner, Misha and Roger, Fabien and others},
  journal={arXiv preprint arXiv:2505.05410},
  year={2025}
}

@article{shao2024deepseekmath,
  title={Deepseekmath: Pushing the limits of mathematical reasoning in open language models},
  author={Shao, Zhihong and Wang, Peiyi and Zhu, Qihao and Xu, Runxin and Song, Junxiao and Bi, Xiao and Zhang, Haowei and Zhang, Mingchuan and Li, YK and Wu, Yang and others},
  journal={arXiv preprint arXiv:2402.03300},
  year={2024}
}

@article{lin2023mitigating,
  title={Mitigating the alignment tax of rlhf},
  author={Lin, Yong and Lin, Hangyu and Xiong, Wei and Diao, Shizhe and Liu, Jianmeng and Zhang, Jipeng and Pan, Rui and Wang, Haoxiang and Hu, Wenbin and Zhang, Hanning and others},
  journal={arXiv preprint arXiv:2309.06256},
  year={2023}
}

@article{cobbe2021training,
  title={Training verifiers to solve math word problems},
  author={Cobbe, Karl and Kosaraju, Vineet and Bavarian, Mohammad and Chen, Mark and Jun, Heewoo and Kaiser, Lukasz and Plappert, Matthias and Tworek, Jerry and Hilton, Jacob and Nakano, Reiichiro and others},
  journal={arXiv preprint arXiv:2110.14168},
  year={2021}
}

@article{jaech2024openai,
  title={Openai o1 system card},
  author={Jaech, Aaron and Kalai, Adam and Lerer, Adam and Richardson, Adam and El-Kishky, Ahmed and Low, Aiden and Helyar, Alec and Madry, Aleksander and Beutel, Alex and Carney, Alex and others},
  journal={arXiv preprint arXiv:2412.16720},
  year={2024}
}

@article{cui2025process,
  title={Process reinforcement through implicit rewards},
  author={Cui, Ganqu and Yuan, Lifan and Wang, Zefan and Wang, Hanbin and Li, Wendi and He, Bingxiang and Fan, Yuchen and Yu, Tianyu and Xu, Qixin and Chen, Weize and others},
  journal={arXiv preprint arXiv:2502.01456},
  year={2025}
}

@article{zha2025rl,
  title={RL Tango: Reinforcing Generator and Verifier Together for Language Reasoning},
  author={Zha, Kaiwen and Gao, Zhengqi and Shen, Maohao and Hong, Zhang-Wei and Boning, Duane S and Katabi, Dina},
  journal={arXiv preprint arXiv:2505.15034},
  year={2025}
}

@article{zou2025reasonflux,
  title={ReasonFlux-PRM: Trajectory-Aware PRMs for Long Chain-of-Thought Reasoning in LLMs},
  author={Zou, Jiaru and Yang, Ling and Gu, Jingwen and Qiu, Jiahao and Shen, Ke and He, Jingrui and Wang, Mengdi},
  journal={arXiv preprint arXiv:2506.18896},
  year={2025}
}

@article{khalifa2025process,
  title={Process reward models that think},
  author={Khalifa, Muhammad and Agarwal, Rishabh and Logeswaran, Lajanugen and Kim, Jaekyeom and Peng, Hao and Lee, Moontae and Lee, Honglak and Wang, Lu},
  journal={arXiv preprint arXiv:2504.16828},
  year={2025}
}

@article{chen2025bridging,
  title={Bridging supervised learning and reinforcement learning in math reasoning},
  author={Chen, Huayu and Zheng, Kaiwen and Zhang, Qinsheng and Cui, Ganqu and Cui, Yin and Ye, Haotian and Lin, Tsung-Yi and Liu, Ming-Yu and Zhu, Jun and Wang, Haoxiang},
  journal={arXiv preprint arXiv:2505.18116},
  year={2025}
}

@article{dong2023raft,
  title={Raft: Reward ranked finetuning for generative foundation model alignment},
  author={Dong, Hanze and Xiong, Wei and Goyal, Deepanshu and Zhang, Yihan and Chow, Winnie and Pan, Rui and Diao, Shizhe and Zhang, Jipeng and Shum, Kashun and Zhang, Tong},
  journal={arXiv preprint arXiv:2304.06767},
  year={2023}
}

@misc{zhao2025genprmscalingtesttimecompute,
      title={GenPRM: Scaling Test-Time Compute of Process Reward Models via Generative Reasoning}, 
      author={Jian Zhao and Runze Liu and Kaiyan Zhang and Zhimu Zhou and Junqi Gao and Dong Li and Jiafei Lyu and Zhouyi Qian and Biqing Qi and Xiu Li and Bowen Zhou},
      year={2025},
      eprint={2504.00891},
      archivePrefix={arXiv},
      primaryClass={cs.CL},
      url={https://arxiv.org/abs/2504.00891}, 
}

@misc{xiong2025stepwiserstepwisegenerativejudges,
      title={StepWiser: Stepwise Generative Judges for Wiser Reasoning}, 
      author={Wei Xiong and Wenting Zhao and Weizhe Yuan and Olga Golovneva and Tong Zhang and Jason Weston and Sainbayar Sukhbaatar},
      year={2025},
      eprint={2508.19229},
      archivePrefix={arXiv},
      primaryClass={cs.AI},
      url={https://arxiv.org/abs/2508.19229}, 
}

@misc{xiong2024implementation,
  title={An implementation of generative prm},
  author={Xiong, Wei and Zhang, Hanning and Jiang, Nan and Zhang, Tong},
  year={2024}
}

@article{eisenstein2023helping,
  title={Helping or herding? reward model ensembles mitigate but do not eliminate reward hacking},
  author={Eisenstein, Jacob and Nagpal, Chirag and Agarwal, Alekh and Beirami, Ahmad and D'Amour, Alex and Dvijotham, DJ and Fisch, Adam and Heller, Katherine and Pfohl, Stephen and Ramachandran, Deepak and others},
  journal={arXiv preprint arXiv:2312.09244},
  year={2023}
}

@article{yang2024qwen2,
  title={Qwen2. 5-math technical report: Toward mathematical expert model via self-improvement},
  author={Yang, An and Zhang, Beichen and Hui, Binyuan and Gao, Bofei and Yu, Bowen and Li, Chengpeng and Liu, Dayiheng and Tu, Jianhong and Zhou, Jingren and Lin, Junyang and others},
  journal={arXiv preprint arXiv:2409.12122},
  year={2024}
}

@article{dubey2024llama,
  title={The llama 3 herd of models},
  author={Dubey, Abhimanyu and Jauhri, Abhinav and Pandey, Abhinav and Kadian, Abhishek and Al-Dahle, Ahmad and Letman, Aiesha and Mathur, Akhil and Schelten, Alan and Yang, Amy and Fan, Angela and others},
  journal={arXiv e-prints},
  pages={arXiv--2407},
  year={2024}
}

@article{xiong2025minimalist,
  title={A minimalist approach to llm reasoning: from rejection sampling to reinforce},
  author={Xiong, Wei and Yao, Jiarui and Xu, Yuhui and Pang, Bo and Wang, Lei and Sahoo, Doyen and Li, Junnan and Jiang, Nan and Zhang, Tong and Xiong, Caiming and others},
  journal={arXiv preprint arXiv:2504.11343},
  year={2025}
}

@article{yuan2024self,
  title={Self-rewarding language models},
  author={Yuan, Weizhe and Pang, Richard Yuanzhe and Cho, Kyunghyun and Sukhbaatar, Sainbayar and Xu, Jing and Weston, Jason},
  journal={arXiv preprint arXiv:2401.10020},
  volume={3},
  year={2024}
}

@article{dong2024rlhf,
  title={Rlhf workflow: From reward modeling to online rlhf},
  author={Dong, Hanze and Xiong, Wei and Pang, Bo and Wang, Haoxiang and Zhao, Han and Zhou, Yingbo and Jiang, Nan and Sahoo, Doyen and Xiong, Caiming and Zhang, Tong},
  journal={arXiv preprint arXiv:2405.07863},
  year={2024}
}

@article{xiong2025self,
  title={Self-rewarding correction for mathematical reasoning},
  author={Xiong, Wei and Zhang, Hanning and Ye, Chenlu and Chen, Lichang and Jiang, Nan and Zhang, Tong},
  journal={arXiv preprint arXiv:2502.19613},
  year={2025}
}

@article{xiong2024building,
  title={Building math agents with multi-turn iterative preference learning},
  author={Xiong, Wei and Shi, Chengshuai and Shen, Jiaming and Rosenberg, Aviv and Qin, Zhen and Calandriello, Daniele and Khalman, Misha and Joshi, Rishabh and Piot, Bilal and Saleh, Mohammad and others},
  journal={arXiv preprint arXiv:2409.02392},
  year={2024}
}

@article{zhang2024policy,
  title={Policy filtration in rlhf to fine-tune llm for code generation},
  author={Zhang, Chuheng and Shen, Wei and Zhao, Li and Zhang, Xuyun and Qi, Lianyong and Dou, Wanchun and Bian, Jiang},
  year={2024}
}

@article{yu2025rip,
  title={Rip: Better models by survival of the fittest prompts},
  author={Yu, Ping and Yuan, Weizhe and Golovneva, Olga and Wu, Tianhao and Sukhbaatar, Sainbayar and Weston, Jason and Xu, Jing},
  journal={arXiv preprint arXiv:2501.18578},
  year={2025}
}

@article{xu2025not,
  title={Not all rollouts are useful: Down-sampling rollouts in llm reinforcement learning},
  author={Xu, Yixuan Even and Savani, Yash and Fang, Fei and Kolter, Zico},
  journal={arXiv preprint arXiv:2504.13818},
  year={2025}
}

@inproceedings{kim2024m,
  title={" I'm Not Sure, But...": Examining the Impact of Large Language Models' Uncertainty Expression on User Reliance and Trust},
  author={Kim, Sunnie SY and Liao, Q Vera and Vorvoreanu, Mihaela and Ballard, Stephanie and Vaughan, Jennifer Wortman},
  booktitle={Proceedings of the 2024 ACM conference on fairness, accountability, and transparency},
  pages={822--835},
  year={2024}
}

@article{yu2025dapo,
  title={Dapo: An open-source llm reinforcement learning system at scale},
  author={Yu, Qiying and Zhang, Zheng and Zhu, Ruofei and Yuan, Yufeng and Zuo, Xiaochen and Yue, Yu and Dai, Weinan and Fan, Tiantian and Liu, Gaohong and Liu, Lingjun and others},
  journal={arXiv preprint arXiv:2503.14476},
  year={2025}
}

@article{zhang2025lessons,
  title={The lessons of developing process reward models in mathematical reasoning},
  author={Zhang, Zhenru and Zheng, Chujie and Wu, Yangzhen and Zhang, Beichen and Lin, Runji and Yu, Bowen and Liu, Dayiheng and Zhou, Jingren and Lin, Junyang},
  journal={arXiv preprint arXiv:2501.07301},
  year={2025}
}

@article{zheng2024processbench,
  title={Processbench: Identifying process errors in mathematical reasoning},
  author={Zheng, Chujie and Zhang, Zhenru and Zhang, Beichen and Lin, Runji and Lu, Keming and Yu, Bowen and Liu, Dayiheng and Zhou, Jingren and Lin, Junyang},
  journal={arXiv preprint arXiv:2412.06559},
  year={2024}
}

@article{schulman2017proximal,
  title={Proximal policy optimization algorithms},
  author={Schulman, John and Wolski, Filip and Dhariwal, Prafulla and Radford, Alec and Klimov, Oleg},
  journal={arXiv preprint arXiv:1707.06347},
  year={2017}
}

@misc{numina_math_7b,
  author = {Edward Beeching and Shengyi Costa Huang and Albert Jiang and Jia Li and Benjamin Lipkin and Zihan Qina and Kashif Rasul and Ziju Shen and Roman Soletskyi and Lewis Tunstall},
  title = {NuminaMath 7B CoT},
  year = {2024},
  publisher = {Numina & Hugging Face},
  journal = {Hugging Face repository},
  howpublished = {\url{https://huggingface.co/AI-MO/NuminaMath-7B-CoT}}
}

@inproceedings{sheng2025hybridflow,
  title={Hybridflow: A flexible and efficient rlhf framework},
  author={Sheng, Guangming and Zhang, Chi and Ye, Zilingfeng and Wu, Xibin and Zhang, Wang and Zhang, Ru and Peng, Yanghua and Lin, Haibin and Wu, Chuan},
  booktitle={Proceedings of the Twentieth European Conference on Computer Systems},
  pages={1279--1297},
  year={2025}
}

@article{hendrycks2021measuring,
  title={Measuring mathematical problem solving with the math dataset},
  author={Hendrycks, Dan and Burns, Collin and Kadavath, Saurav and Arora, Akul and Basart, Steven and Tang, Eric and Song, Dawn and Steinhardt, Jacob},
  journal={arXiv preprint arXiv:2103.03874},
  year={2021}
}

@article{lewkowycz2022solving,
  title={Solving quantitative reasoning problems with language models},
  author={Lewkowycz, Aitor and Andreassen, Anders and Dohan, David and Dyer, Ethan and Michalewski, Henryk and Ramasesh, Vinay and Slone, Ambrose and Anil, Cem and Schlag, Imanol and Gutman-Solo, Theo and others},
  journal={Advances in neural information processing systems},
  volume={35},
  pages={3843--3857},
  year={2022}
}

@article{he2024olympiadbench,
  title={Olympiadbench: A challenging benchmark for promoting agi with olympiad-level bilingual multimodal scientific problems},
  author={He, Chaoqun and Luo, Renjie and Bai, Yuzhuo and Hu, Shengding and Thai, Zhen Leng and Shen, Junhao and Hu, Jinyi and Han, Xu and Huang, Yujie and Zhang, Yuxiang and others},
  journal={arXiv preprint arXiv:2402.14008},
  year={2024}
}

@article{wang2023math,
  title={Math-shepherd: Verify and reinforce llms step-by-step without human annotations},
  author={Wang, Peiyi and Li, Lei and Shao, Zhihong and Xu, RX and Dai, Damai and Li, Yifei and Chen, Deli and Wu, Yu and Sui, Zhifang},
  journal={arXiv preprint arXiv:2312.08935},
  year={2023}
}

@misc{luo2024improvemathematicalreasoninglanguage,
      title={Improve Mathematical Reasoning in Language Models by Automated Process Supervision}, 
      author={Liangchen Luo and Yinxiao Liu and Rosanne Liu and Samrat Phatale and Meiqi Guo and Harsh Lara and Yunxuan Li and Lei Shu and Yun Zhu and Lei Meng and Jiao Sun and Abhinav Rastogi},
      year={2024},
      eprint={2406.06592},
      archivePrefix={arXiv},
      primaryClass={cs.CL},
      url={https://arxiv.org/abs/2406.06592}, 
}

@article{michaud2020understanding,
  title={Understanding learned reward functions},
  author={Michaud, Eric J and Gleave, Adam and Russell, Stuart},
  journal={arXiv preprint arXiv:2012.05862},
  year={2020}
}

@article{tien2022causal,
  title={Causal confusion and reward misidentification in preference-based reward learning},
  author={Tien, Jeremy and He, Jerry Zhi-Yang and Erickson, Zackory and Dragan, Anca D and Brown, Daniel S},
  journal={arXiv preprint arXiv:2204.06601},
  year={2022}
}

@article{bradley1952rank,
  title={Rank analysis of incomplete block designs: I. the method of paired comparisons},
  author={Bradley, Ralph Allan and Terry, Milton E},
  journal={Biometrika},
  volume={39},
  number={3/4},
  pages={324--345},
  year={1952},
  publisher={JSTOR}
}

@inproceedings{lightman2023let,
  title={Let's verify step by step},
  author={Lightman, Hunter and Kosaraju, Vineet and Burda, Yuri and Edwards, Harrison and Baker, Bowen and Lee, Teddy and Leike, Jan and Schulman, John and Sutskever, Ilya and Cobbe, Karl},
  booktitle={The Twelfth International Conference on Learning Representations},
  year={2023}
}

@article{zhou2023webarena,
  title={Webarena: A realistic web environment for building autonomous agents},
  author={Zhou, Shuyan and Xu, Frank F and Zhu, Hao and Zhou, Xuhui and Lo, Robert and Sridhar, Abishek and Cheng, Xianyi and Ou, Tianyue and Bisk, Yonatan and Fried, Daniel and others},
  journal={arXiv preprint arXiv:2307.13854},
  year={2023}
}

@article{jimenez2023swe,
  title={Swe-bench: Can language models resolve real-world github issues?},
  author={Jimenez, Carlos E and Yang, John and Wettig, Alexander and Yao, Shunyu and Pei, Kexin and Press, Ofir and Narasimhan, Karthik},
  journal={arXiv preprint arXiv:2310.06770},
  year={2023}
}

@article{zhu2025chain,
  title={Chain-of-thought matters: improving long-context language models with reasoning path supervision},
  author={Zhu, Dawei and Wei, Xiyu and Zhao, Guangxiang and Wu, Wenhao and Zou, Haosheng and Ran, Junfeng and Wang, Xun and Sun, Lin and Zhang, Xiangzheng and Li, Sujian},
  journal={arXiv preprint arXiv:2502.20790},
  year={2025}
}

@inproceedings{lyu2023faithful,
  title={Faithful chain-of-thought reasoning},
  author={Lyu, Qing and Havaldar, Shreya and Stein, Adam and Zhang, Li and Rao, Delip and Wong, Eric and Apidianaki, Marianna and Callison-Burch, Chris},
  booktitle={The 13th International Joint Conference on Natural Language Processing and the 3rd Conference of the Asia-Pacific Chapter of the Association for Computational Linguistics (IJCNLP-AACL 2023)},
  year={2023}
}

@inproceedings{yeo2024interpretable,
  title={How interpretable are reasoning explanations from prompting large language models?},
  author={Yeo, Wei Jie and Satapathy, Ranjan and Goh, Rick Siow Mong and Cambria, Erik},
  booktitle={Findings of the Association for Computational Linguistics: NAACL 2024},
  year={2024}
}

@inproceedings{turpin2023language,
  title={Language Models Don't Always Say What They Think: Unfaithful Explanations in Chain-of-Thought Prompting},
  author={Turpin, Miles and Michael, Julian and Perez, Ethan and Bowman, Samuel R.},
  booktitle={Advances in Neural Information Processing Systems 36 (NeurIPS 2023)},
  year={2023}
}

@inproceedings{nguyen2024direct,
  title={Direct Evaluation of Chain-of-Thought in Multi-hop Reasoning with Knowledge Graphs},
  author={Nguyen, Minh-Vuong and Luo, Linhao and Shiri, Fatemeh and Phung, Dinh and Li, Yuan-Fang and Vu, Thuy-Trang and Haffari, Gholamreza},
  booktitle={Findings of the Association for Computational Linguistics: ACL 2024},
  year={2024}
}

@inproceedings{paul2024making,
  title={Making Reasoning Matter: Measuring and Improving Faithfulness of Chain-of-Thought Reasoning},
  author={Paul, Debjit and West, Robert and Bosselut, Antoine and Faltings, Boi},
  booktitle={Findings of the Association for Computational Linguistics: EMNLP 2024},
  year={2024}
}

@inproceedings{jacovi2024chain,
  title={A Chain-of-Thought Is as Strong as Its Weakest Link: A Benchmark for Verifiers of Reasoning Chains},
  author={Jacovi, Alon and Bitton, Yonatan and Bohnet, Bernd and Herzig, Jonathan and Honovich, Or and Tseng, Michael and Collins, Michael and Aharoni, Roee and Geva, Mor},
  booktitle={Proceedings of the 62nd Annual Meeting of the Association for Computational Linguistics (Volume 1: Long Papers)},
  year={2024}
}
